\documentclass[sigconf]{acmart}

\usepackage{hyperref}
\hypersetup{hidelinks,
	colorlinks=true,
	allcolors=black,
	pdfstartview=Fit,
	breaklinks=true}

\usepackage{algorithm}
\usepackage{algorithmic}
\usepackage{amsmath}
\usepackage{amssymb}
\usepackage{color}
\usepackage{natbib}
\usepackage{multirow}
\usepackage{subfigure}
\usepackage{bbding}

\AtBeginDocument{%
  \providecommand\BibTeX{{%
    \normalfont B\kern-0.5em{\scshape i\kern-0.25em b}\kern-0.8em\TeX}}}

\setcopyright{acmcopyright}
\copyrightyear{2023}
\acmYear{2023}
\acmDOI{XXXXXXX.XXXXXXX}

\acmConference[KDD '23]{29th ACM SIGKDD Conference on Knowledge Discovery and Data Mining}{August 06--10, 2023}{Long Beach, CA}
\acmPrice{15.00}
\acmISBN{978-1-4503-XXXX-X/23/08}

\begin{document}

\title{Complementary Classifier Induced Partial Label Learning}

\author{Yuheng Jia}
\authornote{Both authors contributed equally to this research.}
\authornote{Corresponding author.}
\affiliation{%
  \institution{Key Laboratory of Computer Network and Information Integration, School of Computer Science and Engineering}
  \streetaddress{Southeast University}
  \city{Nanjing 210096}
  \country{China}}
\email{yhjia@seu.edu.cn}

\author{Chongjie Si}
\authornotemark[1]
\affiliation{%
  \institution{MoE Key Lab of Artificial Intelligence, AI Institute}
  \streetaddress{Shanghai Jiao Tong University}
  \city{Shanghai 200240}
  \country{China}}
\email{chongjiesi@sjtu.edu.cn}

\author{Min-Ling Zhang}
\affiliation{%
  \institution{Key Laboratory of Computer Network and Information Integration, School of Computer Science and Engineering}
  \streetaddress{Southeast University}
  \city{Nanjing 210096}
  \country{China}
}
\email{zhangml@seu.edu.cn}

\renewcommand{\shortauthors}{Trovato and Tobin, et al.}

\begin{abstract}
 In partial label learning (PLL), each training sample is associated with a set of candidate labels, among which only one is valid. The core of PLL is to disambiguate the candidate labels to get the ground-truth one. In disambiguation, the existing works usually do not fully investigate the effectiveness of the non-candidate label set (a.k.a. complementary labels), which accurately indicates a set of labels that do not belong to a sample. In this paper, we use the non-candidate labels to induce a complementary classifier, which naturally forms an adversarial relationship against the traditional PLL classifier, to eliminate the false-positive labels in the candidate label set. Besides, we assume the feature space and the label space share the same local topological structure captured by a dynamic graph, and use it to assist disambiguation. Extensive experimental results validate the superiority of the proposed approach against state-of-the-art PLL methods on 4 controlled UCI data sets and 6 real-world data sets, and reveal the usefulness of complementary learning in PLL. \textit{The code has been released in the link \url{https://github.com/Chongjie-Si/PL-CL}.}

\end{abstract}



\maketitle

\section{Introduction}
In traditional multi-class learning, each training sample is associated with a single label indicating the class that sample belongs to. However, collecting such highly accurate label is expensive and time-consuming, which is also a bottleneck of many real-world classification tasks. Therefore, partial label learning (PLL) \cite{hullermeier2006learningPL-KNN}, \cite{zhang2016partialPLLEAF}, \cite{lyu2019gmshdjdfjd} was proposed, where in PLL, an instance is associated with a set of candidate labels, among which only one is valid. PLL has been applied to many real-world applications, such as natural language processing \cite{yang2014semidsdgsg}, web mining \cite{huiskes2008mirMirflickr} and image classification \cite{zeng2013learningSoccerplayer}, \cite{liu2012conditional23462372}, etc.

Formally speaking, suppose $\mathcal{X}=\mathbb{R}^q$ denotes the $q$-dimensional feature space, $\mathcal{Y} =\{0,1\}^l$ is the label space with $l$ classes. Given a partial label data set $\mathcal{D} = \{\mathbf{x}_i, S_i | 1\leq i \leq n\}$ where $\mathbf{x}_i \in \mathcal{X}$ is the $i$-th sample, and $S_i \subseteq \mathcal{Y}$ is the corresponding candidate label set and $n$ is the number of samples, the task of PLL is to learn a function $f: \mathcal{X}\rightarrow \mathcal{Y}$ based on $\mathcal{D}$. As the ground-truth labels $\{y_1,y_2,...,y_n\}$ are concealed in the corresponding candidate label sets, i.e., $y_i \in S_i$, which are not directly accessible, PLL is a quite challenging problem.

Accordingly, disambiguating the candidate labels becomes the core task of PLL. A naive strategy is to treat all the candidate labels of a sample equally, and then averages the modeling outputs of the candidate labels, which is known as the averaging-based framework \cite{cour2011learningsssnyturtues}. For example, PL-KNN \cite{hullermeier2006learningPL-KNN} averages the candidate labels of neighboring instances to make prediction. However, this naive strategy may fail when false-positive labels dominate the candidate label set. The second strategy is to identify the ground-truth label from the candidate label set, which is known as the identification-based framework (a.k.a. the disambiguation-based approach).
\begin{figure*}
    \centering
    \includegraphics[scale=0.5]{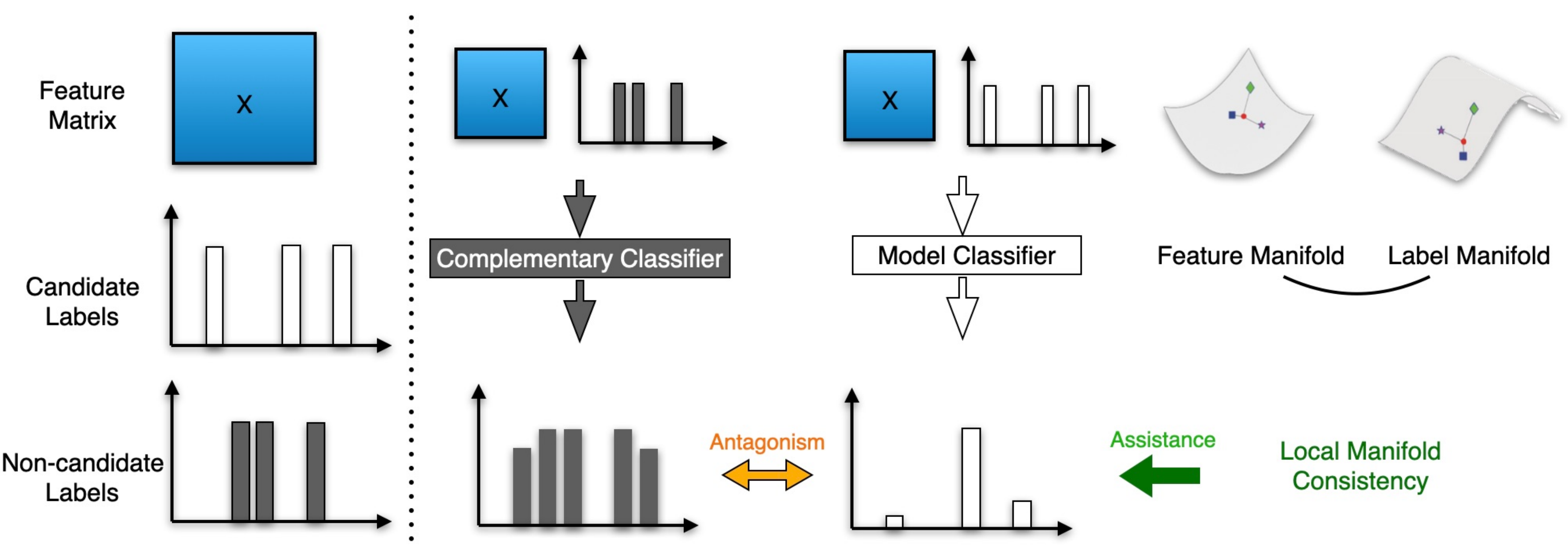}
     \caption{Framework of PL-CL. PL-CL utilizes candidate and non-candidate labels to generate an ordinary classifier and a complementary classifier, and restrains the outputs of those two classifiers with an adversarial manner. A dynamic graph is also adopted to exploit the local topology consistency of the feature manifold and label manifold to further assist disambiguation.}
    \label{fig:framework}
\end{figure*}
To disambiguate the candidate labels, we believe the following three kinds of prior information are important, and the previous methods usually use one or two kinds of priors to achieve disambiguation. The first type is the correlations among instances, i.e., if the two samples are similar to each other in the feature space, those two samples are possible to have the same label. For instance, PL-AGGD \cite{wang2021adaptivePLAGGD} and IPAL \cite{zhang2015solvingIPAL} use the feature matrix to construct a similarity graph and further use that graph to realize disambiguation. The second kind of prior is the mapping from instances to the candidate labels, i.e., we could expect that the mapping error of an instance to the correct label is small, while the mapping error of that instance with an incorrect label in the candidate label set (i.e., false-positive label) is relatively large. For example, SURE \cite{2019PartialSURE} adopts such a mapping to automatically remove the incorrect labels by maximizing an infinity norm.  The last kind of information is non-candidate label set, which precisely describes ``a set of labels should not be assigned to a sample". Although the information in non-candidate label set is directly available and accurate, the majority of existing works overlook the valuable information in disambiguation. 

To fully take advantage of above mentioned priors, as shown in Fig. \ref{fig:framework}, we propose a novel PLL method named PL-CL, with \textbf{P}artial \textbf{L}abel learning based on \textbf{C}omplementary \textbf{L}earning. Specifically, PL-CL uses the candidate labels to design an ordinary classifier, and meanwhile adopts the non-candidate labels to develop a complementary classifier. As the ordinary classifier indicates which label should be assigned to a sample while the complementary classifier specifies the labels that should not be assigned to that sample, the outputs of those two classifiers form an adversarial relationship. Moreover, as the non-candidate labels are accurate, the trained complementary classifier is highly reliable and can be used to promote the ordinary classifier through the adversarial relationship, which finally enhances PLL. Additionally, inspired by \cite{hou2016multimanifold}, the feature space and label space of a data set usually lie in two different manifolds but own the similar local structure. Therefore, PL-CL constructs an adaptive local topological graph shared by both the feature space and the label space. Then similar samples in the feature space will possess the similar labels, which further benefits the disambiguation of PLL. PL-CL is finally formulated as an adaptive graph regularized dual classifiers learning problem, and solved by an alternating iterative algorithm. Comprehensive experiments on both the controlled UCI data sets and the real-world data sets validate that PL-CL performs significantly better than state-of-the-art PLL methods, and also substantiate the effectiveness of the complementary classifier and the shared adaptive local graph structure in PLL.

The rest of the paper is organized as follows. We first review some related works in Section 2 and introduce the proposed approach in Section 3. The numerical solution is provided in Section 4. Experimental results and analyses are shown in Section 5, and conclusions and future research directions are finally given at Section 6.

\section{Related Work}
\subsection{Partial Label Learning}
PLL is a representative weakly supervised learning framework, which learns from ambiguous supervision information. In PLL, an instance is associated with a set of labels, among which only one is valid. Since the ground-truth label of each sample is concealed in its candidate labels, which can not be directly accessible during the training phase, PLL is a quite challenging task.

To address the mentioned challenge, the key approach is candidate label disambiguation, i.e., \cite{2018LeveragingLALO}, \cite{nguyen2008classificationplsvm}, \cite{2019PartialSURE}, \cite{zhang2015solvingIPAL}. The existing methods can be summarized into two categories: averaging-based approaches and identification-based approaches. For the averaging-based approaches \cite{hullermeier2006learningPL-KNN}, 
\cite{cour2011learningsssnyturtues}, each candidate label of a training sample is treated in an equal manner and the model prediction is yielded by averaging the modeling outputs of the candidate labels. For example, PL-KNN \cite{hullermeier2006learningPL-KNN} averages the candidate labels of neighboring instances to make prediction. This strategy is intuitive, however, it may fail when false-positive labels dominate the candidate label set. For the identification-based approaches, \cite{wang2021adaptivePLAGGD}, \cite{2019PartialSURE}, \cite{2018LeveragingLALO}, the valid label is considered as a latent variable and can be identified through an iterative optimization procedure. For instance, PL-AGGD \cite{wang2021adaptivePLAGGD} uses the feature matrix to construct a similarity graph, which is further utilized for generating labeling confidence of each candidate label.

In this paper, we summarize the following three kinds of priors that are important for disambiguation. The first is type is the correlations among instances, i.e., similar samples in the feature space may share a same label. The second kind of prior is the mapping from instances to the candidate labels, i.e., the mapping error of an instance to the valid label is small, while that to an incorrect label in the candidate label set is large. The last kind of information is non-candidate label set, which precisely describes a set of labels should not be assigned to a sample. Existing works mainly focus on the first two types of priors, while the valuable information in non-candidate labels (complementary labels) which is directly available and accurate is overlooked. Therefore, we will fully take advantage of above mentioned priors (especially the complementary labels) to achieve PLL.

\subsection{Learning from Complementary Labels}

Complementary-label learning (CLL) is another representative framework of weakly supervised learning \cite{feng2020learning}, \cite{yu2018learning}. In the traditional CLL \cite{ishida2017learning}, an instance is associated with only one label, which indicates that this sample does not belong to. Although in this paper, complementary label information are adopted, it is quite different from that in the traditional complementary-label learning. First, in partial label learning the number of complementary labels of each sample is usually more than one, while an instance only has one complementary label in complementary-label learning. Additionally, the aim of complementary classifier in PLL is to assist the disambiguation of candidate labels, which is different from CLL. To the best of our knowledge, the information of complementary labels has been overlooked to some extent in previous PLL works, and our work is the first time to induce a complementary classifier to assist the construction of the ordinary classifier by an adversarial prior in PLL.


\section{The Proposed Approach}
PLL aims to induce a multi-class classifier $f: \mathcal{X}\rightarrow \mathcal{Y}$ based on the data set $\mathcal{D} = \{\mathbf{x}_i, S_i | 1\leq i \leq n\}$. Denote $\mathbf{X} = [\mathbf{x}_1,\mathbf{x}_2,...,\mathbf{x}_n]^{\mathsf{T}} \in \mathbb{R}^{n\times q}$ the instance matrix and $\mathbf{Y} = [\mathbf{y}_1,\mathbf{y}_2,...,\mathbf{y}_n]^{\mathsf{T}} \in \{0,1\}^{n\times l}$ the partial label matrix with $l$ labels. Note that the value of $y_{ij}$ is binary, and $y_{ij} = 1$ (resp. $y_{ij}=0$) if the $j$-th label resides in (resp. does not reside in) the candidate label set of $\mathbf{x}_i$. 

As analyzed in the introduction, three kinds of priors are important to label disambiguation. We use all the three kinds of priors to solve the PLL problem, and formulate the loss function as 
\begin{equation}
    \min  \mathcal{L}(\mathbf{X},\mathbf{Y})+\hat{\mathcal{L}}(\mathbf{X}, \hat{\mathbf{Y}}) + \Psi(\mathbf{X},\mathbf{Y}), 
\end{equation}
where $\hat{\mathbf{Y}}=[\hat{y}_{ij}]_{n\times l}$ is the complementary label matrix, i.e., $\hat{y}_{ij}=1$ (resp. $\hat{y}_{ij}=0$) if the $j$-th label is in the non-candidate (resp. candidate) label set of the $i$-th instance. $\mathcal{L}(\mathbf{X},\mathbf{Y})$ denotes the mapping from instance matrix to the candidate labels, i.e., the ordinary classifier. $\hat{\mathcal{L}}(\mathbf{X},\mathbf{Y})$ denotes the loss of the complementary classifier, i.e., the mapping from the instance to the non-candidate labels, and $\Psi(\mathbf{X},\mathbf{Y})$ uses correlations of instances to disambiguate the candidate labels. These three terms are three different ways of disambiguation in PLL, and details of them will be introduced in the following subsections.

\subsection{Ordinary Classifier}
First, PL-CL uses a weight matrix $\mathbf{W} =[\mathbf{w}_1, \mathbf{w}_2,...,\mathbf{w}_l] \in \mathbb{R}^{q \times l}$ to map the instance matrix to the labeling confidence matrix $\mathbf{P}=[\mathbf{p}_1,\mathbf{p}_2,...,\mathbf{p}_n]^\mathsf{T} \in \mathbb{R}^{n\times l}$, where $\mathbf{p}_i \in \mathbb{R}^l$ is the labeling confidence vector corresponding to $\mathbf{x}_i$, and $p_{ij}$ represents the probability of the $j$-th label being the ground-truth label of $\mathbf{x}_i$. To be specific, PL-CL uses the following least squares loss to form the ordinary classifier:

\begin{equation}
\begin{split}
   \mathcal{L}(\mathbf{X}, \mathbf{Y})&= \left\|\mathbf{XW}+\mathbf{1}_n\mathbf{b}^\mathsf{T} - \mathbf{P} \right\|_F^2 + \lambda\|\mathbf{W}\|_F^2\\
   {\rm s.t.} &\quad \sum_{j}p_{ij} = 1, 0\leq p_{ij} \leq y_{ij}, \forall i, j, 
   \end{split}
    \label{eq basic model}
\end{equation}
where $\mathbf{b} = [b_1,b_2,...,b_l]\in \mathbb{R}^l$ is the bias term and $\mathbf{1}_n \in \mathbb{R}^n$ is an all ones vector. $\|\mathbf{W}\|_F$ denotes the Frobenius norm of the weight matrix, and minimizing $\|\mathbf{W}\|^2_F$ will control the complexity of the model. $\lambda$ is a hyper-parameter trading off these terms. As $\mathbf{P}$ represents the labeling confidence, it should be strictly non-negative, i.e., $p_{ij}\geq0$
and $p_{ij}$ has a chance to be positive only when the $j$-th label lies in the candidate label set of $i$-th sample, i.e., we have $p_{ij}\leq y_{ij}$. Moreover, as each row of $\mathbf{P}$ indicates the labeling confidence of a certain sample, the sum of its elements should be equal to 1, i.e., $\sum_j p_{ij}=1$. The ideal state of $\mathbf{p}_i$ is one-hot, for there will only one ground-truth label. By minimizing Eq. (\ref{eq basic model}), the ground-truth label is excepted to produce a smaller mapping error than the incorrect ones residing the candidate label set, and the correct label is likely to be selected.

\subsection{Complementary Classifier}
Suppose matrix $\mathbf{Q}=[\mathbf{q}_1,\mathbf{q}_2,...,\mathbf{q}_n]\in \mathbb{R}^{n\times l}$ is the complementary-labeling confidence matrix where $q_{ij}$ denotes the confidence of the $j$-th label NOT being the ground-truth label of $\mathbf{x}_i$. Similar as the ordinary classifier, we construct a complementary classifier:

\begin{equation}
\begin{split}
    \min_{\hat{\mathbf{W}},\hat{\mathbf{b}} } & \left\|\mathbf{X}\hat{\mathbf{W}} + \mathbf{1}_n\hat{\mathbf{b}}^{\mathsf{T}} - \mathbf{Q}\right\|_F^2+\lambda \left\|\hat{\mathbf{W}}\right\|_F^2 \\
     {\rm s.t.} & \quad\hat{y}_{ij}\leq q_{ij} \leq 1, \forall i, j, 
    \end{split}
\end{equation}
 where $\hat{\mathbf{W}}\in \mathbb{R}^{q\times l}$ and $\hat{\mathbf{b}}\in\mathbb{R}^l$ are the weight matrix and bias for the complementary classifier. Different from $\mathbf{p}_i$, the ideal state of $\mathbf{q}_i$ is ``zero-hot'', i.e., only one value in $\mathbf{q}_i$ is 0 with others 1, for there is only one ground-truth label in PLL. Therefore, we require $\hat{y}_{ij}\leq q_{ij}\leq 1, \forall i,j$, which means $0\leq q_{ij}\leq 1$ if the $j$-th label resides in the candidate label set, and $q_{ij}=1$, if $\hat{y}_{ij}=1$.
 
 The ordinary classifier predicts which label should be assigned to an instance while the complementary classifier specifies a group of labels that should not be allocated to that instance. \textit{Therefore, an adversarial relationship exists between the outputs of the ordinary and complementary classifiers, i.e., a larger (resp. smaller) element in $\mathbf{P}$ implies a smaller (resp. larger) element in $\mathbf{Q}$, and vice versa.} Besides, as the elements in both $\mathbf{Q}$ and $\mathbf{P}$ are non-negative and no more than 1, the above adversarial relationship  can be formulated as  $\mathbf{p}_i +\mathbf{q}_i = \mathbf{1}_l $. Taking the adversarial relationship into account, the complementary classifier becomes:

\begin{equation}
\begin{split}
     \hat{\mathcal{L}}(\mathbf{X}, \hat{\mathbf{Y}}) =&  \alpha\left\|\mathbf{X}\hat{\mathbf{W}} + \mathbf{1}_n\hat{\mathbf{b}}^{\mathsf{T}} - \mathbf{Q}\right\|_F^2 \\ & +\beta 
     \left\|\mathbf{1}_{n\times l} - \mathbf{P}-\mathbf{Q}\right\|_F^2 +\lambda \left\|\hat{\mathbf{W}}\right\|_F^2\\
     {\rm s.t.} & \quad\hat{y}_{ij}\leq q_{ij} \leq 1,
\end{split}
    \label{eq complementary learning}
\end{equation}
where $\mathbf{1}_{n\times l}\in\mathbb{R}^{n\times l}$ is an all ones matrix, and $\alpha$ and $\beta$ are two hyper-parameters. Different from the candidate label set, the complementary information in the non-candidate label set is directly accessible and accurate, by minimizing Eq. (\ref{eq complementary learning}), the valuable information is passed to the labeling confidence matrix $\mathbf{P}$, which will further assist label disambiguation.

\subsection{Local Consistency in Label Manifold and Feature Manifold}
As suggested by \cite{hou2016multimanifold}, the label space and feature space of a data set lie in two different manifolds but with the similar local structure, i.e., if $\mathbf{x}_i$ and $\mathbf{x}_j$ are similar, their corresponding labeling confidence vector $\mathbf{p}_i$ and $\mathbf{p}_j$ should also be similar. To this end, we first construct a local non-negative similarity matrix $\mathbf{G}=[g_{ij}]_{n\times n}$ to capture the local manifold structure of the feature space, i.e.,

\begin{equation}
\begin{split}
        \min_{\mathbf{G}} &\quad  \left\|\mathbf{X}^\mathsf{T}-\mathbf{X}^\mathsf{T}\mathbf{G}\right\|_F^2 \\
        {\rm s.t.} &\quad \mathbf{G}^\mathsf{T}\mathbf{1}_n = \mathbf{1}_n, \mathbf{0}_{n\times n} \leq \mathbf{G} \leq \mathbf{U},
\end{split}
\label{eq graph G}
\end{equation}
where $\mathbf{0}_{n\times n}$ is an all zeros matrix with the size of $n\times n$. $\mathbf{G}^\mathsf{T}\mathbf{1}_n=\mathbf{1}_n$ will ensure that the sum of each column of $\mathbf{G}$ is 1, i.e.,  $\mathbf{G}$ is normalized. $\mathbf{U} = [u_{ij}]_{n\times n}$ denotes the $k$-nearest neighbors (KNN) location matrix, i.e., $u_{ij} = 1$ if $\mathbf{x}_i$ belongs to the KNN of $\mathbf{x}_j$, otherwise $u_{ij} = 0$. $\mathbf{0}_{n\times n} \leq \mathbf{G} \leq \mathbf{U}$ means that the elements of $\mathbf{G}$ will be no less than 0 and no more than the elements of $\mathbf{U}$ at the same location, which will ensure that one sample can only be reconstructed by its neighbors. We further assume the label space shares the same local structure as the feature space captured by $\mathbf{G}$, and use it to assist label disambiguation, i.e.,

\begin{equation}
\begin{split}
        \min_{\mathbf{P}} &\quad  \left\|\mathbf{P}^\mathsf{T}-\mathbf{P}^\mathsf{T}\mathbf{G}\right\|_F^2 \\
        {\rm s.t.} &\quad \mathbf{P}\mathbf{1}_q = \mathbf{1}_n, \mathbf{0}_{n\times l} \leq \mathbf{P} \leq \mathbf{Y}. 
\end{split}
\label{eq graph P}
\end{equation}
By minimizing Eq. (\ref{eq graph P}), the local manifold structure $\mathbf{G}$ will help produce a better $\mathbf{P}$. Finally, we combine Eqs. (\ref{eq graph G}) and (\ref{eq graph P}) together to jointly optimize the local structure $\mathbf{G}$ and the labeling confidence matrix $\mathbf{P}$, i.e.,

\begin{equation}
    \begin{split}
        \Psi (\mathbf{X},\mathbf{Y})= &\quad \gamma \left\|\mathbf{X}^\mathsf{T}-\mathbf{X}^\mathsf{T}\mathbf{G}\right\|_F^2 + \mu \left\|\mathbf{P}^\mathsf{T}-\mathbf{P}^\mathsf{T}\mathbf{G}\right\|_F^2 \\
        {\rm s.t.} &\quad\mathbf{G}^\mathsf{T}\mathbf{1}_n = \mathbf{1}_n, \mathbf{0}_{n\times n} \leq \mathbf{G} \leq \mathbf{U},\\
        &\quad \mathbf{P}\mathbf{1}_q = \mathbf{1}_n, \mathbf{0}_{n\times l} \leq \mathbf{P} \leq \mathbf{Y},
    \end{split}
    \label{eq Graph definition}
\end{equation}
where $\gamma$ and $\mu$ are two hyper-parameters. By minimizing Eq. (\ref{eq Graph definition}), the feature space and label space will be well communicated to achieve a better local consistency.

\subsection{Model Formulation}
Based on the discussions above, the objective function of PL-CL is finally formulated as

\begin{equation}
    \begin{split}
        \min_{\substack{\mathbf{W},\mathbf{b},\hat{\mathbf{W}},\hat{\mathbf{b}},\\ \mathbf{P},\mathbf{Q},\mathbf{G}}} &  \left\|\mathbf{XW}+\mathbf{1}_n\mathbf{b}^\mathsf{T} - \mathbf{P} \right\|_F^2 + \beta\|\mathbf{1}_{n\times l} - \mathbf{P}-\mathbf{Q}||_F^2 \\
        +&\alpha \left\|\mathbf{X}\hat{\mathbf{W}} + \mathbf{1}_n\hat{\mathbf{b}}^{\mathsf{T}} - \mathbf{Q}\right\|_F^2+\mu \left\|\mathbf{P}^\mathsf{T}-\mathbf{P}^\mathsf{T}\mathbf{G}\right\|_F^2 \\
        +&\gamma \left\|\mathbf{X}^\mathsf{T}-\mathbf{X}^\mathsf{T}\mathbf{G}\right\|_F^2 + \lambda (\|{{\mathbf{W}}}\|_F^2+ \left\|{\hat{\mathbf{W}}}\right\|_F^2) \vspace{2pt}\\
        {\rm s.t.}\quad  & \mathbf{P}\mathbf{1}_q = \mathbf{1}_n, \mathbf{0}_{n\times l} \leq \mathbf{P} \leq \mathbf{Y}, \hat{\mathbf{Y}}\leq\mathbf{Q}\leq \mathbf{1}_{n\times l} \\
         &\mathbf{G}^\mathsf{T}\mathbf{1}_n = \mathbf{1}_n, \mathbf{0}_{n\times n} \leq \mathbf{G} \leq \mathbf{U},
        \end{split}
    \label{eq objective function}
\end{equation}
where $\beta,\alpha,\gamma,\mu$ and $\lambda$ are the hyper-parameters balancing different terms. PL-CL comprehensively takes the information in the candidate-label set, the non-candidate label set, and the sample relationships into consideration to perform label disambiguation. As will be shown later, it is quite robust to all these hyper-parameters.

The most challenging issue in PLL is the inaccurate supervision. As the non-candidate labels precisely indicate whether an instance does not belong to a certain class, which can be regarded as a kind of accurate supervision, leveraging the non-candidate labels will certainly help PLL. 

To the best of our knowledge, it is the first time to use the non-candidate labels to construct a complementary classifier in PLL to help label disambiguation. Moreover, the adversarial prior to introduce the complementary classifier is also novel in PLL.

\section{Numerical Solution}
The optimization problem in Eq. (\ref{eq objective function}) has seven variables with different constraints, we therefore adopt an alternative and iterative manner to solve it.

\subsection{Update W, b and $\hat{\mathbf{W}}$, $\hat{\mathbf{b}}$}

With other variables fixed, problem (\ref{eq objective function}) with respect to $\mathbf{W}$ and $\mathbf{b}$ can be reformulated as

\begin{equation}
    \min_{\mathbf{W}, \mathbf{b}}\quad   \left\|\mathbf{XW}+\mathbf{1}_n\mathbf{b}^\mathsf{T} - \mathbf{P} \right\|_F^2 + \lambda  \left\|\mathbf{W}\right\|_F^2, 
    \label{eq W b problem}
\end{equation}
which is a regularized least squares problem and the corresponding closed-form solution is 

\begin{equation}
    \begin{split}
        \mathbf{W} &= \left(\mathbf{X}^\mathsf{T}\mathbf{X}+\lambda\mathbf{I}_{n\times n} \right)^{-1}\mathbf{X}^\mathsf{T}\mathbf{P}\\
        \mathbf{b}&=\frac{1}{n}\left(\mathbf{P}^\mathsf{T}\mathbf{1}_n - \mathbf{W}^\mathsf{T}\mathbf{X}^\mathsf{T}\mathbf{1}_n \right),
    \end{split}
\end{equation}
where $\mathbf{I}_{n\times n}$ is an identity matrix with the size of $n\times n$. Similarly, the problem (\ref{eq objective function}) with respect to $\hat{\mathbf{W}}$ and $\hat{\mathbf{b}}$ is

\begin{equation}
    \min_{\hat{\mathbf{W}}, \hat{\mathbf{b}}}\quad  \|\mathbf{X}\hat{\mathbf{W}} + \mathbf{1}_n\hat{\mathbf{b}}^{\mathsf{T}} - \mathbf{Q}\|_F^2 + \frac{\lambda}{\alpha} \|\hat{\mathbf{W}}\|_F^2,
\end{equation}
which has a similar analytical solution as Eq. (\ref{eq W b problem}), i.e.,

\begin{equation}
    \begin{split}
        \hat{\mathbf{W}} &= \left(\mathbf{X}^\mathsf{T}\mathbf{X}+\frac{\lambda}{\alpha}\mathbf{I}_{n\times n}\right)^{-1}\mathbf{X}^\mathsf{T}\mathbf{Q}\\
        \hat{\mathbf{b}}&=\frac{1}{n}\left(\mathbf{Q}^\mathsf{T}\mathbf{1}_n - \hat{\mathbf{W}}^\mathsf{T}\mathbf{X}^\mathsf{T}\mathbf{1}_n\right).
    \end{split}
\end{equation}

\noindent
\textbf{Kernel Extension} The above linear model may be too simple to tackle the complex relationships between the instances to labels, therefore we extend the linear model to a kernel version. Suppose $\phi(\cdot):\mathbb{R}^q \rightarrow \mathbb{R}^h$ represents a feature transformation that maps the feature space to a higher dimensional space, and $\mathbf{\Phi}= [\phi(\mathbf{x}_1), \phi(\mathbf{x}_2),...,\phi(\mathbf{x}_n) ]$ denotes the feature matrix in the higher dimensional space. With the feature mapping, we can rewrite problem (\ref{eq W b problem}) as follows:

\begin{equation}
\begin{split}
    \min_{\mathbf{W}, \mathbf{b}}\quad& \|\mathbf{M} \|_F^2 + \lambda \|\mathbf{W}\|_F^2\\
    {\rm s.t.}\quad& \mathbf{M} = \mathbf{\Phi}\mathbf{W}+\mathbf{1}_n\mathbf{b}^\mathsf{T} - \mathbf{P},
    \end{split}
    \label{eq A b problem}
\end{equation}
where $\mathbf{M} = [\mathbf{m}_1,\mathbf{m}_2,...,\mathbf{m}_n ] \in \mathbb{R}^{n \times l}$. The Lagrangian function of Eq. (\ref{eq A b problem}) is formulated as 

\begin{equation}
\begin{split}
    \mathcal{L}(\mathbf{W},\mathbf{b},\mathbf{M},\mathbf{A}) = \|\mathbf{M} \|_F^2 + \lambda \|\mathbf{W}\|_F^2 \\- {\rm tr}\left(\mathbf{A}^{\mathsf{T}}(\mathbf{\Phi}\mathbf{W}+\mathbf{1}_n\mathbf{b}^\mathsf{T} - \mathbf{P}-\mathbf{M})\right),
\end{split}
\end{equation}
where $\mathbf{A} \in \mathbb{R}^{n\times l}$ is the Lagrange multiplier. ${\rm tr}(\cdot)$ is the trace of a matrix. Based on the KKT conditions, we have

\begin{equation}
    \begin{split}
        &\frac{\partial \mathcal{L}}{\partial \mathbf{W}} = 2\lambda\mathbf{W} -\mathbf{\Phi}^{\mathsf{T}}\mathbf{A}=0, \frac{\partial \mathcal{L}}{\partial \mathbf{b}} = \mathbf{A}^{\mathsf{T}}\mathbf{1}_n=0 \\
       & \frac{\partial \mathcal{L}}{\partial \mathbf{M}} = 2\mathbf{M}+ \mathbf{A}=0,   \frac{\partial \mathcal{L}}{\partial \mathbf{A}} = \mathbf{\Phi}\mathbf{W}+\mathbf{1}_n\mathbf{b}^\mathsf{T} - \mathbf{P}-\mathbf{M}=0.
       \label{eq A b gradient}
    \end{split}
\end{equation}
Define a kernel matrix $\mathbf{K}=\mathbf{\Phi}\mathbf{\Phi}^{\mathsf{T}}$ with its element $k_{ij} = \phi(\mathbf{x}_i)\phi(\mathbf{x}_j) = \mathcal{K}(\mathbf{x}_i,\mathbf{x}_j)$, where $\mathcal{K}(\cdot,\cdot)$ is the kernel function. For PL-CL, we use Gaussian function $\mathcal{K}(\mathbf{x}_i,\mathbf{x}_j) = {\rm exp}(-\|\mathbf{x}_i - \mathbf{x}_j\|_2^2/(2\sigma^2))$ as the kernel function and set $\sigma$ to the average distance of all pairs of training instances. Then, we have

\begin{equation}
\begin{split}
    \mathbf{A} &= \left(\frac{1}{2\lambda} \mathbf{K} + \frac{1}{2}\mathbf{I}_{n\times n}\right)^{-1}\left(\mathbf{P} - \mathbf{1}_n\mathbf{b}^{\mathsf{T}}\right),\\
     \mathbf{b} &= \left(\frac{\mathbf{sP}}{\mathbf{s1}_n}\right)^{\mathsf{T}},
\end{split}
    \label{eq A b solution}
\end{equation}
where $\mathbf{s} =\mathbf{1}_n^{\mathsf{T}}\left(\frac{1}{2\lambda} \mathbf{K} + \frac{1}{2}\mathbf{I}_{n\times n}\right)^{-1}$. We can also obtain $\mathbf{W} = {\mathbf{\Phi}^{\mathsf{T}} \mathbf{A}}/{2\lambda}$ from Eq. (\ref{eq A b gradient}). The output of the model can be denoted by $\mathbf{H} = \mathbf{\Phi}\mathbf{W} +\mathbf{1}_n\mathbf{b}^{\mathsf{T}} = \frac{1}{2\lambda}\mathbf{KA} + \mathbf{1}_n\mathbf{b}^{\mathsf{T}}$.

As the optimization problem regarding $\hat{\mathbf{W}}$ and $\hat{\mathbf{b}}$ has the same form as problem (\ref{eq A b problem}), similarly, we have
\begin{equation}
\begin{split}
    \hat{\mathbf{A}} &= \left(\frac{\alpha}{2\lambda} \mathbf{K} + \frac{1}{2}\mathbf{I}_{n\times n}\right)^{-1}\left(\mathbf{Q} - \mathbf{1}_n\hat{\mathbf{b}}^{\mathsf{T}}\right),\\
    \hat{\mathbf{b}} &= \left(\frac{\hat{\mathbf{s}}\mathbf{Q}}{\hat{\mathbf{s}}\mathbf{1}_n}\right)^{\mathsf{T}}, 
\end{split}
\label{eq what bhat solution}
\end{equation}
where $\hat{\mathbf{s}} =\mathbf{1}_n^{\mathsf{T}}\left(\frac{\alpha}{2\lambda} \mathbf{K} + \frac{1}{2}\mathbf{I}_{n\times n}\right)^{-1}$, and the output of the complementary classifier is defined as $\hat{\mathbf{H}} = \frac{\alpha}{2\lambda}\mathbf{K}\hat{\mathbf{A}} + \mathbf{1}_n\hat{\mathbf{b}}^{\mathsf{T}}$.

\subsection{Update Q}
Fixing other variables, the $\mathbf{Q}$-subproblem can be equivalently rewritten as

\begin{equation}
    \begin{split}
    \min_{\mathbf{Q}}\quad& \left\|\mathbf{Q} - \frac{\alpha\hat{\mathbf{H}} + \beta(\mathbf{1}_{n\times l}-\mathbf{P})}{\alpha+\beta}\right\|_F^2 \\
    {\rm s.t.}\quad & \hat{\mathbf{Y}}\leq\mathbf{Q}\leq \mathbf{1}_{n\times l},
\end{split}
\label{eq Q problem reformat}
\end{equation}
where the solution is 

\begin{equation}
    \mathbf{Q} =  \mathcal{T}_1\left(\mathcal{T}_{\hat{\mathbf{Y}}}\left(\frac{\alpha\hat{\mathbf{H}} + \beta(\mathbf{1}_{n\times l}-\mathbf{P})}{\alpha+\beta}\right)\right).
    \label{eq Q solution}
\end{equation}
$\mathcal{T}_1$, $\mathcal{T}_{\hat{\mathbf{Y}}}$ are two thresholding operators in element-wise, i.e., $\mathcal{T}_1(m):=\min\{1,m\}$ with $m$ being a scalar and $\mathcal{T}_{\hat{\mathbf{Y}}}(m):=\max\{\hat{{y}}_{ij}, m\}$.

\subsection{Update G}

Fixing other variables, the $\mathbf{G}$-subproblem is rewritten as 

\begin{equation}
\begin{split}
    \min_{\mathbf{G}} \quad & \gamma\left\|\mathbf{X}^\mathsf{T}-\mathbf{X}^\mathsf{T}\mathbf{G}\right\|_F^2 + \mu\left\|\mathbf{P}^\mathsf{T}-\mathbf{P}^\mathsf{T}\mathbf{G}\right\|_F^2\\
    {\rm s.t.}\quad & \mathbf{G}^\mathsf{T}\mathbf{1}_n = \mathbf{1}_n, \mathbf{0}_{n\times n} \leq \mathbf{G} \leq \mathbf{U}.
\end{split}
\label{eq G problem}
\end{equation}
Notice that each column of $\mathbf{G}$ is independent to other columns, therefore we can solve the $\mathbf{G}$-subproblem column by column. To solve the $i$-th column of $\mathbf{G}$, we have

\begin{equation}
\begin{split}
    \min_{\mathbf{G_{\cdot i}}} \quad & \gamma\left\|\mathbf{x}_i - \sum_{k_{ji}=1} g_{ji}\mathbf{x}_j \right\|_F^2 + \mu\left\|\mathbf{p}_i-\sum_{k_{ji}=1} g_{ji}\mathbf{p}_j\right\|_F^2\\
    {\rm s.t.}\quad & \mathbf{G}_{\cdot i}^\mathsf{T}\mathbf{1}_n = 1, \mathbf{0}_{n} \leq \mathbf{G}_{\cdot i} \leq \mathbf{U}_{\cdot i}.
\end{split}
\label{eq G subproblem}
\end{equation}
As there are only $k$ non-zero elements in $\mathbf{G}_{\cdot i}$ that are supposed to be updated, which corresponds to the reconstruction coefficients of $\mathbf{x}_i$ by its top-$k$ neighbors, let $\hat{\mathbf{g}}_i\in \mathbb{R}^k$ denote the vector of these elements. Let $\mathcal{N}_i$ be the set of the indexes of these neighbors corresponding to $\hat{\mathbf{g}}_i$. Define matrix $\mathbf{D}^{x_i} = [\mathbf{x}_i-\mathbf{x}_{\mathcal{N}_{i(1)}}, \mathbf{x}_i-\mathbf{x}_{\mathcal{N}_{i(2)}},...,\mathbf{x}_i-\mathbf{x}_{\mathcal{N}_{i(k)}}]^\mathsf{T} \in \mathbb{R}^{k\times l}$ and $\mathbf{D}^{p_i} = [\mathbf{p}_i-\mathbf{p}_{\mathcal{N}_{i(1)}}, \mathbf{p}_i-\mathbf{p}_{\mathcal{N}_{i(2)}},...,\mathbf{p}_i-\mathbf{p}_{\mathcal{N}_{i(k)}}]^\mathsf{T} \in \mathbb{R}^{k\times l}$, let Gram matrices $\mathbf{B}^{x_i} = \mathbf{D}^{x_i}(\mathbf{D}^{x_i})^\mathsf{T} \in \mathbb{R}^{k\times k}$ and $\mathbf{B}^{p_i} = \mathbf{D}^{p_i}(\mathbf{D}^{p_i})^\mathsf{T} \in \mathbb{R}^{k\times k}$, we can transform the problem in Eq. (\ref{eq G subproblem}) into the following form:

\begin{equation}
    \begin{split}
    \min_{\hat{\mathbf{g}}_i} \quad & \hat{\mathbf{g}}_i^\mathsf{T}\left(\gamma \mathbf{B}^{x_i} + \mu \mathbf{B}^{p_i} \right)\hat{\mathbf{g}}_i\\
    {\rm s.t.}\quad & \hat{\mathbf{g}}_i^\mathsf{T}\mathbf{1}_k = 1, \mathbf{0}_{k} \leq \hat{\mathbf{g}}_i \leq \mathbf{1}_k.
    \end{split}
    \label{eq G solution}
\end{equation}
The optimization problem in Eq. (\ref{eq G solution}) is a standard Quadratic Programming (QP) problem, and can be solved by off-the-shelf QP tools. $\mathbf{G}$ is updated by concatenating all the solved $\hat{\mathbf{g}}_i$ together.

\subsection{Update P}
With other variables fixed, the $\mathbf{P}$-subproblem is 

\begin{equation}
    \begin{split}
    \min_{{\mathbf{P}}} \quad & \left\|\mathbf{H}-\mathbf{P}\right\|_F^2 + \beta\left\|\mathbf{E}-\mathbf{Q}-\mathbf{P}\right\|_F^2+ \mu\left\|\mathbf{P}^\mathsf{T}-\mathbf{P}^\mathsf{T}\mathbf{G}\right\|_F^2\\
    {\rm s.t.}\quad & \mathbf{P}\mathbf{1}_q = \mathbf{1}_n, \mathbf{0}_{n\times l} \leq \mathbf{P} \leq \mathbf{Y},
    \end{split}
    \label{eq P problem}
\end{equation}
and we rewrite the problem in Eq. (\ref{eq P problem})
into the following form:

\begin{equation}
    \begin{split}
    \min_{{\mathbf{P}}} \quad & \left\|\mathbf{P}-\frac{\mathbf{H}+\beta(\mathbf{E}-\mathbf{Q})}{1+\beta}\right\|_F^2 + \frac{\mu}{1+\beta} \left \|\mathbf{P}^\mathsf{T}-\mathbf{P}^\mathsf{T}\mathbf{G} \right \|_F^2\\
    {\rm s.t.}\quad & \mathbf{P}\mathbf{1}_q = \mathbf{1}_n, \mathbf{0}_{n\times l} \leq \mathbf{P} \leq \mathbf{Y}.
    \end{split}
    \label{eq P rewritten problem}
\end{equation}
In order to solve the problem (\ref{eq P rewritten problem}), we denote $\widetilde{\mathbf{p}} = {\rm vec}\left(\mathbf{P}\right) \in [0,1]^{nl}$, where ${\rm vec}(\cdot)$ is the vectorization operator. Similarly, $\widetilde{\mathbf{o}} = {\rm vec}\left(\frac{\mathbf{H}+\beta(\mathbf{E}-\mathbf{Q})}{1+\beta} \right) \in \mathbb{R}^{nl}$ and $\widetilde{\mathbf{y}} = {\rm vec}(\mathbf{Y}) \in\{0,1\}^{nl}$. Let $\mathbf{T} = 2(\mathbf{I}_{n\times n}-\mathbf{G})(\mathbf{I}_{n\times n}-\mathbf{G})^\mathsf{T} \in \mathbb{R}^{n \times n}$ be a square matrix. Based on these notations, the optimization problem (\ref{eq P rewritten problem}) can be written as 

\begin{equation}
    \begin{split}
    \min_{\widetilde{\mathbf{p}}} \quad & \frac{1}{2} \widetilde{\mathbf{p}}^\mathsf{T} \left( \mathbf{C} + \frac{2(1+\beta)}{\mu}\mathbf{I}_{nl\times nl} \right)\widetilde{\mathbf{p}}-\frac{2(1+\beta)}{\mu}\widetilde{\mathbf{o}}^\mathsf{T}\widetilde{\mathbf{p}}\\
    {\rm s.t.}\quad & \sum_{i=1,i\% n = j, 0\leq j\leq n-1}^{nl}\widetilde{\mathbf{p}}_i=1, \mathbf{0}_{nl} \leq \widetilde{\mathbf{p}}\leq \widetilde{\mathbf{y}},
    \end{split}
    \label{eq P solution}
\end{equation}
where $\mathbf{C}\in \mathbb{R}^{nl\times nl}$ is defined as:

\begin{equation}
   {\mathbf{C}}=\left[\begin{array}{cccc}
{\mathbf{T}} & \mathbf{0}_{m \times m} & \cdots & \mathbf{0}_{m \times m} \\
\mathbf{0}_{m \times m} & {\mathbf{T}} & \ddots & \vdots \\
\vdots & \ddots & \ddots & \mathbf{0}_{m \times m} \\
\mathbf{0}_{m \times m} & \cdots & \mathbf{0}_{m \times m} & {\mathbf{T}}
\end{array}\right].
\end{equation}
As the problem in Eq. (\ref{eq P solution}) is a standard QP problem, it can be solved by off-the-shelf QP tools.

\subsection{The Overall Optimization Algorithm}
As a summary, PL-CL first initializes  $\mathbf{G}$ and $\mathbf{P}$ by solving problems (\ref{eq graph G}) and (\ref{eq graph P}) respectively, and initializes $\mathbf{Q}$ as:

\begin{equation}
q_{i j}=\left\{\begin{array}{ll}
\frac{1}{l-\sum_{j} y_{i j}} & \text { if } \quad y_{i j}=0 \\
0 & \text { otherwise }.
\end{array}\right.
\label{eq Q ini}
\end{equation}
Then, PL-CL iteratively and alternatively updates one variable with the other fixed until the model converges. For an unseen instance $\mathbf{x}^\ast$, the predicted label $y^\ast$ is 

\begin{equation}
    y^{*} = \mathop{\arg\max}_{k} \sum_{i=1}^n \frac{1}{2\lambda}\mathbf{A}_{ik}\mathcal{K}(\mathbf{x}^\ast, \mathbf{x}_i)+b_k.
    \label{eq prediction}
\end{equation}
The pseudo code of PL-CL is summarized in Algorithm \ref{alg:PLCL}.

\begin{algorithm}[tb]
   \caption{The pseudo code of PL-CL}
   \label{alg:PLCL}
\begin{algorithmic}
   \STATE {\bfseries Input:} Partial label data $\mathcal{D}$, hyper-parameter $\alpha$, $\beta$, $\mu$, $\gamma$ and $\lambda$, an unseen sample $\mathbf{x}^\ast$
   \STATE {\bfseries Output:} The predicted label $y^\ast$ of $\mathbf{x}^\ast$
   \STATE {\bfseries Process}:
   \STATE Initialize $\mathbf{G}$ according to Eq. (\ref{eq graph G}).
   \STATE Initialize $\mathbf{P}$ according to Eq. (\ref{eq graph P}).
   \STATE Initialize $\mathbf{Q}$ according to Eq. (\ref{eq Q ini}).
   \REPEAT
   \STATE Update $\mathbf{A}$, $\mathbf{b}$ according to Eq. (\ref{eq A b solution}).
   \STATE Update $\hat{\mathbf{A}}$, $\hat{\mathbf{b}}$ according to Eq. (\ref{eq what bhat solution}).
   \STATE Update $\mathbf{Q}$ according to Eq. (\ref{eq Q solution}).
   \STATE Update $\mathbf{G}$ according to Eq. (\ref{eq G solution}).
   \STATE Update $\mathbf{P}$ according to Eq. (\ref{eq P solution}).

   \UNTIL{Convergence}
   \STATE {\bfseries Return}: $y^\ast$ according to Eq. (\ref{eq prediction}).
\end{algorithmic}
\end{algorithm}

\subsection{Complexity Analysis}

The complexity for solving $\mathbf{W}$, $\mathbf{b}$, $\hat{\mathbf{W}}$ and $\hat{\mathbf{b}}$ is $O(n^3) + O(nql)$, that for updating $\mathbf{Q}$ is $O(nl)$, and complexities for updating $\mathbf{G}$ and $\mathbf{P}$ are $O(nk^3)$ and $O(n^3q^3)$. Therefor the overall complexity of PL-CL is $O(n^3+nql+nl+nk^3+n^3q^3)$.

\section{Experiments}
\subsection{Compared Methods}

In order to validate the effectiveness of PL-CL, we compared it with six state-of-the-art PLL approaches: 

\begin{itemize}
    \item PL-AGGD \cite{wang2021adaptivePLAGGD}: a PLL approach which constructs a similarity graph and further uses that graph to realize disambiguation [hyper-parameter configurations: $k = 10$, $T = 20$, $\lambda = 1$, $\mu = 1$, $\gamma = 0.05$];
    \item SURE \cite{2019PartialSURE}: a self-guided retraining based approach which maximizes an infinity norm regularization on the modeling outputs to realize disambiguation [hyper-parameter configurations: $\lambda$ and $\beta$ $\in \{0.001, 0.01, 0.05, 0.1, 0.3, 0.5, 1\}$]; 
    \item LALO \cite{2018LeveragingLALO}: an identification-based approach with constrained local consistency to differentiate the candidate labels [hyper-parameter configurations: $k=10$, $\lambda=0.05$, $\mu=0.005$];
    \item IPAL \cite{zhang2015solvingIPAL}: an instance-based algorithm which identifies the valid label via an iterative label propagation procedure [hyper-parameter configurations: $\alpha = 0.95$, $k = 10$, $T = 100$];
    \item PLDA \cite{wang2022partialPLDA}: a dimensionality reduction approach for partial label learning [hyper-parameter configurations: PL-KNN];
    \item PL-KNN \cite{hullermeier2006learningPL-KNN}: an averaging-based method that averages the neighbors' candidate labels to make prediction on an unseen sample [hyper-parameter configurations: $k=10$].
\end{itemize}

\subsection{Experimental Settings}
For PL-CL, we set $k=10$, $\lambda =0.03$ , $\gamma$, $\mu$, $\alpha$ and $\beta$ were chosen from $\{0.001,0.01,0.1,0.2,0.5,1,1.5,2,4\}$. The kernel functions of SURE, PL-AGGD and LALO were the same as that of PL-CL. All the hyper-parameter configurations of each method were set according to the original paper. Ten runs of 50\%/50\% random train/test splits were performed, and the averaged accuracy with standard deviations on test data sets were recorded for all the comparing algorithms.

\begin{table*}[ht]

\renewcommand\arraystretch{0.8} 
    \centering
    \caption{Characteristics of the real-world data sets.}

    \begin{tabular}{|c c c c c c|}
     \hline \hline
        Data set & \# Examples & \# Features & \# Labels & \# Average Labels & Task Domain\\\hline
      
        FG-NET \cite{panis2016overviewFg-net} & 1002 & 262 & 78 & 7.48 & facial age estimation\\
         Lost \cite{cour2009learningLost} & 1122 & 108 & 16 & 2.23 & automatic face naming\\
        MSRCv2 \cite{liu2012conditionalMSRCv2} & 1758 & 48 & 23 & 3.16 & object classification\\
        Mirflickr \cite{huiskes2008mirMirflickr}& 2780 & 1536 & 14 & 2.76 & web image classification \\
        Soccer Player \cite{zeng2013learningSoccerplayer}& 17472 & 279 & 171 & 2.09 & automatic face naming\\
        Yahoo!News \cite{guillaumin2010multipleYahoonews} & 22991 & 163 & 219 & 1.91 & automatic face naming \\
        \hline\hline
    \end{tabular}
    \label{tab:real-world data character}
\end{table*}

\begin{table*}[ht!]\footnotesize

\setlength{\tabcolsep}{1.8mm}
\renewcommand\arraystretch{1} 
    \centering
    \caption{The comparison of different methods on the real-world data sets. $\bullet$/$\circ$ indicates whether PL-CL is statistically superior/inferior to the compared algorithm according to pairwise $t$-test at significance level of 0.05.}
    \begin{tabular}{|c l l l l l l l l|}\hline\hline
         Data set & \multicolumn{1}{c}{FG-NET} & \multicolumn{1}{c}{Lost} & \multicolumn{1}{c}{MSRCv2} & \multicolumn{1}{c}{Mirflickr} & \multicolumn{1}{c}{Soccer Player} &
         \multicolumn{1}{c}{Yahoo!News}&
         \multicolumn{1}{c}{FG-NET(MAE3)}&
         \multicolumn{1}{c|}{FG-NET(MAE5)}\\\hline
         
         PL-CL & 0.072 $\pm$ 0.009 & 0.709 $\pm$ 0.022  & 0.469 $\pm$ 0.016 & 0.642 $\pm$ 0.012 & 0.534 $\pm$ 0.004 & 0.618 $\pm$ 0.003 & 0.433 $\pm$ 0.022 & 0.575 $\pm$ 0.015 \\
         SURE &  0.052 $\pm$ 0.006 $\bullet$ & 0.693 $\pm$ 0.020 $\bullet$ & 0.445 $\pm$ 0.021 $\bullet$ & 0.631 $\pm$ 0.021 $\bullet$ & 0.519 $\pm$ 0.004 $\bullet$ & 0.598 $\pm$ 0.002 $\bullet$ &  0.356 $\pm$ 0.019 $\bullet$ & 0.494 $\pm$ 0.020 $\bullet$ \\
         PL-AGGD & 0.063 $\pm$ 0.009 $\bullet$ & 0.683 $\pm$ 0.014 $\bullet$ & 0.451 $\pm$ 0.012 $\bullet$ & 0.610 $\pm$ 0.011 $\bullet$ & 0.524 $\pm$ 0.004 $\bullet$ & 0.607 $\pm$ 0.004 $\bullet$ & 0.387 $\pm$ 0.015 $\bullet$ & 0.530 $\pm$ 0.015 $\bullet$ \\ 
         LALO & 0.065 $\pm$ 0.009 $\bullet$ & 0.680 $\pm$ 0.014 $\bullet$ & 0.448 $\pm$ 0.015 $\bullet$ & 0.626 $\pm$ 0.013 $\bullet$ & 0.523 $\pm$ 0.003 $\bullet$ & 0.600 $\pm$ 0.003 $\bullet$ & 0.423 $\pm$ 0.020 $\bullet$ & 0.566 $\pm$ 0.014 $\bullet$ \\
         IPAL & 0.051 $\pm$ 0.011 $\bullet$ & 0.646 $\pm$ 0.023 $\bullet$ & 0.488 $\pm$ 0.031 $\circ$ & 0.527 $\pm$ 0.009 $\bullet$ & 0.528 $\pm$ 0.003 $\bullet$ & 0.625 $\pm$ 0.004 $\circ$ &  0.349 $\pm$ 0.019 $\bullet$ & 0.500 $\pm$ 0.019 $\bullet$ \\
         PLDA & 0.042 $\pm$ 0.005 $\bullet$ & 0.289 $\pm$ 0.045 $\bullet$ & 0.422 $\pm$ 0.013 $\bullet$ & 0.480 $\pm$ 0.015 $\bullet$ & 0.493 $\pm$ 0.003 $\bullet$ & 0.380 $\pm$ 0.003 $\bullet$ & 0.150 $\pm$ 0.012 $\bullet$ & 0.232 $\pm$ 0.012 $\bullet$ \\
         PL-KNN & 0.036 $\pm$ 0.006 $\bullet$ & 0.296 $\pm$ 0.021 $ \bullet$ & 0.393 $\pm$ 0.014 $\bullet$ & 0.454 $\pm$ 0.015 $\bullet$ & 0.483 $\pm$ 0.005 $\bullet$ & 0.368 $\pm$ 0.004 $\bullet$ & 0.288 $\pm$ 0.013 $\bullet$ & 0.440 $\pm$ 0.016 $\bullet$ \\\hline\hline
    \end{tabular}
     
    \label{tab:real world statistic}
\end{table*}

\begin{table}[ht]
\setlength{\tabcolsep}{2mm}
\renewcommand\arraystretch{0.8} 
    \centering
    \caption{Characteristics of the UCI data sets.}
 
    \begin{tabular}{|c c c c c|}
     \hline \hline
        Data set & ecoli & glass & vehicle & steel \\\hline
        \# Examples & 336 & 214 & 846 & 1941\\
        \# Features & 7 & 9 & 18 & 27\\
        \# Labels &8 & 6 & 4 & 7\\
        \hline\hline
    \end{tabular}
    \label{tab:uci data character}
    
\end{table}

\begin{table}[ht]
\setlength{\tabcolsep}{1.5mm}
\renewcommand\arraystretch{1} 
    \centering
    \caption{Win/tie/loss counts on the controlled UCI data sets between PL-CL and other compared approaches according to the pairwise $t$-test at 0.05 significance level. I: varying $\epsilon$; II: varying $p$ ($r=1$); III: varying $p$ ($r=2$); IV: varying $p$ ($r=3$).}
    \begin{tabular}{|c c c c c c|}
     \hline \hline
        & I & II  & III & IV & Total \\\hline
        SURE & 22/6/0 & 26/2/0 & 26/2/0 & 24/4/0 & 88/14/0\\
        PL-AGGD & 26/2/0 & 25/3/0 & 23/5/0 & 22/6/0 & 96/16/0\\
        LALO & 27/1/0 & 24/4/0 & 26/2/0 & 22/6/0 & 99/13/0\\
        IPAL & 28/0/0 & 23/5/0 & 26/2/0 & 25/3/0 & 102/10/0\\
        PLDA & 28/0/0 & 28/0/0 & 28/0/0 & 28/0/0 & 112/0/0 \\
        PL-KNN & 28/0/0 & 28/0/0 & 28/0/0 & 28/0/0 & 112/0/0\\
        \hline\hline
    \end{tabular}
   
    \label{tab:uci data wintieloss}
    
\end{table}

\begin{figure*} [ht!]
	\centering
	
	\includegraphics[scale=1]{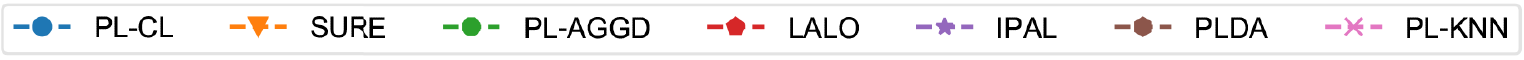}\\
	\subfigure[Classification accuracy on controlled UCI data sets with $\epsilon$ varying from 0.1 to 0.7 ($p=1$, $r=1$).]{
		\includegraphics[scale=0.263]{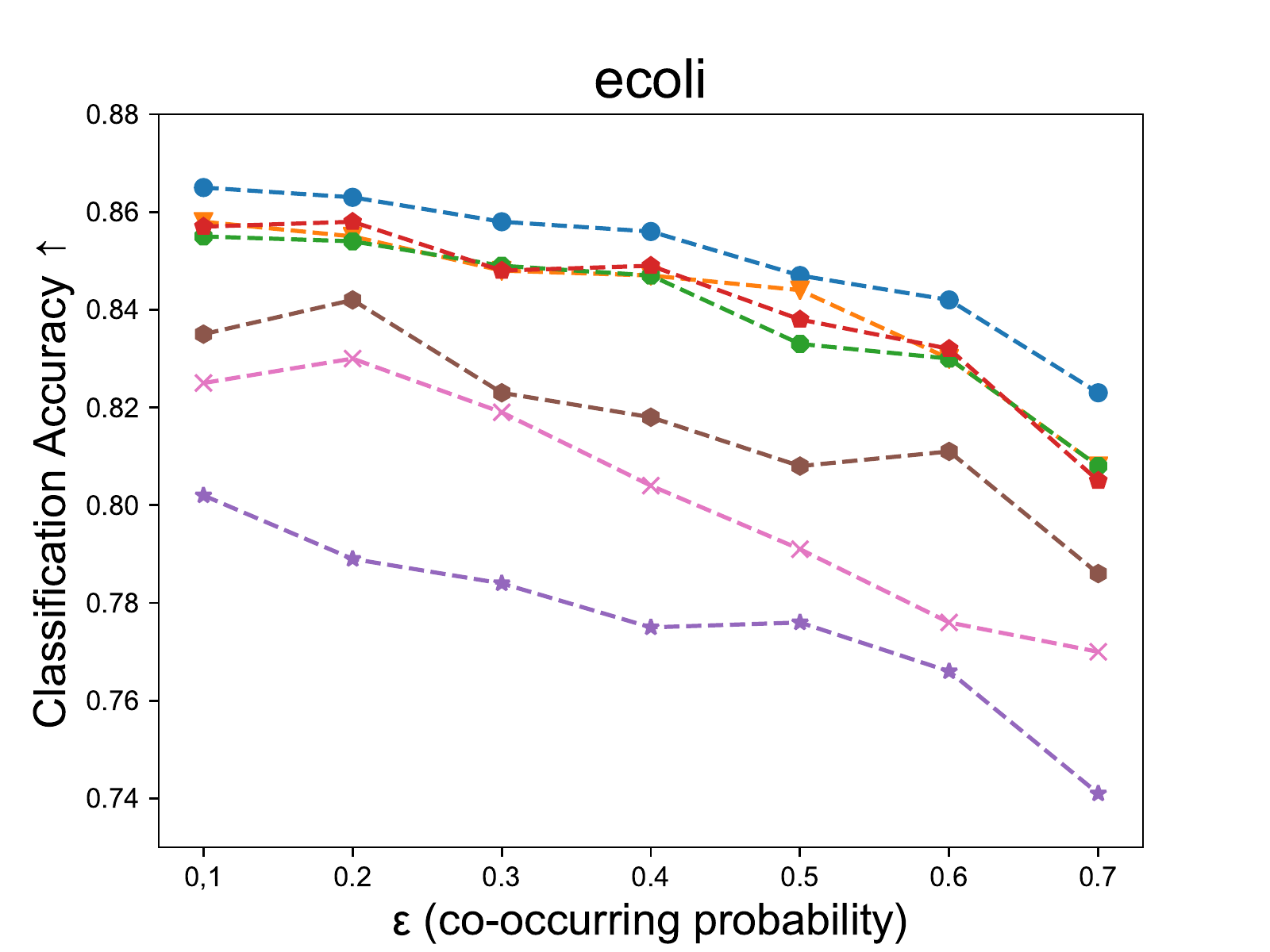}
		\includegraphics[scale=0.263]{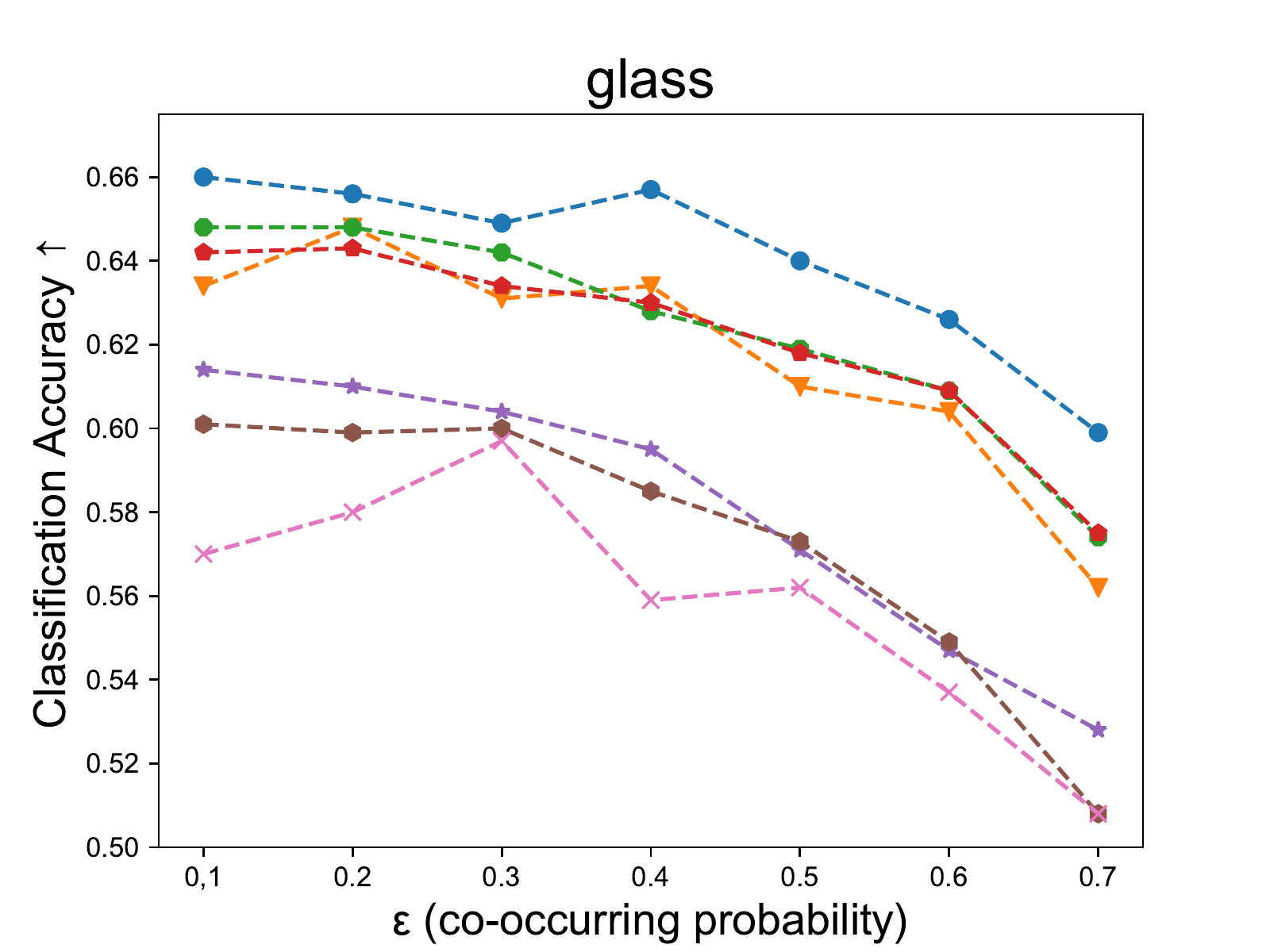}
		\includegraphics[scale=0.263]{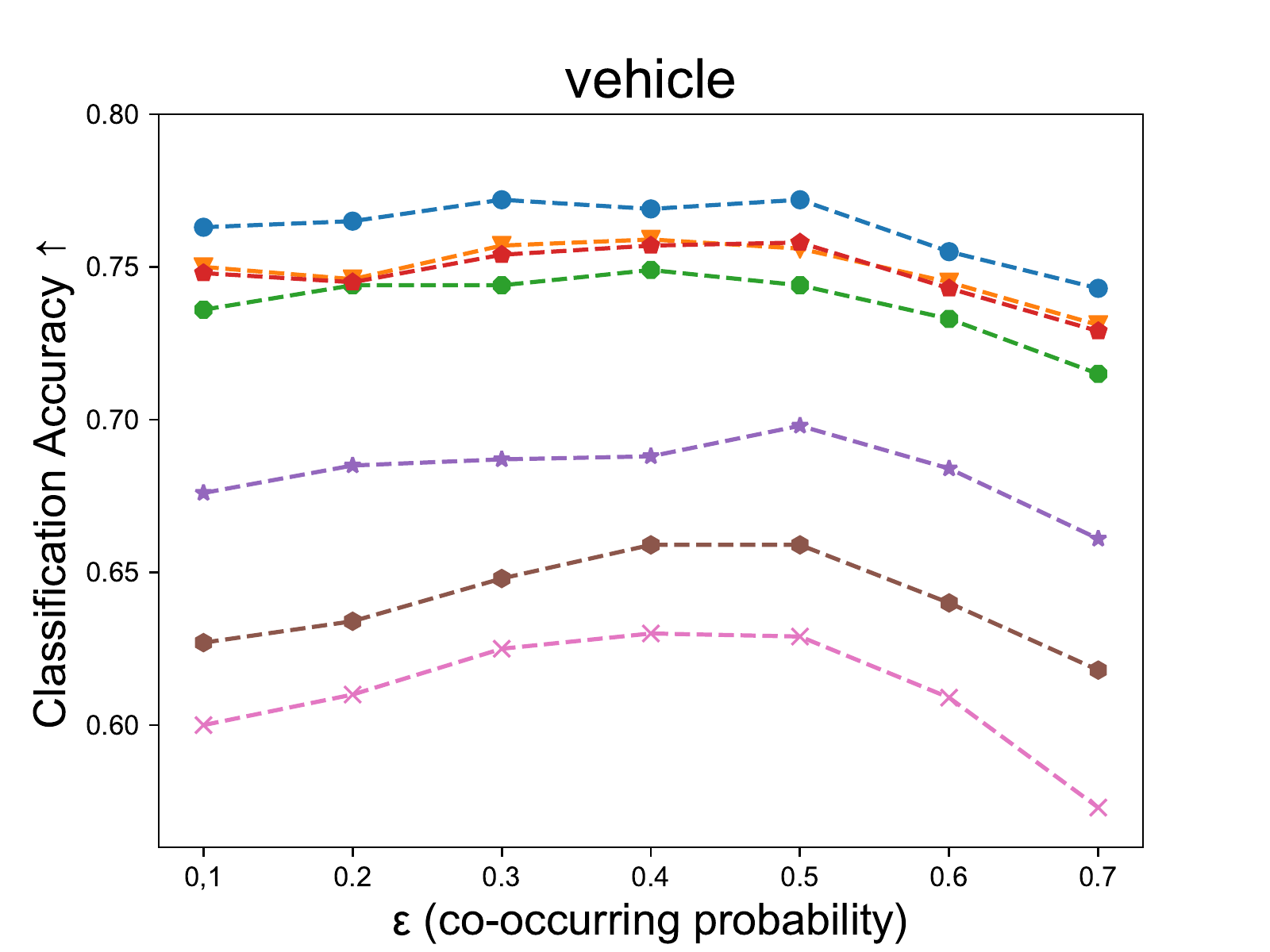}
		\includegraphics[scale=0.263]{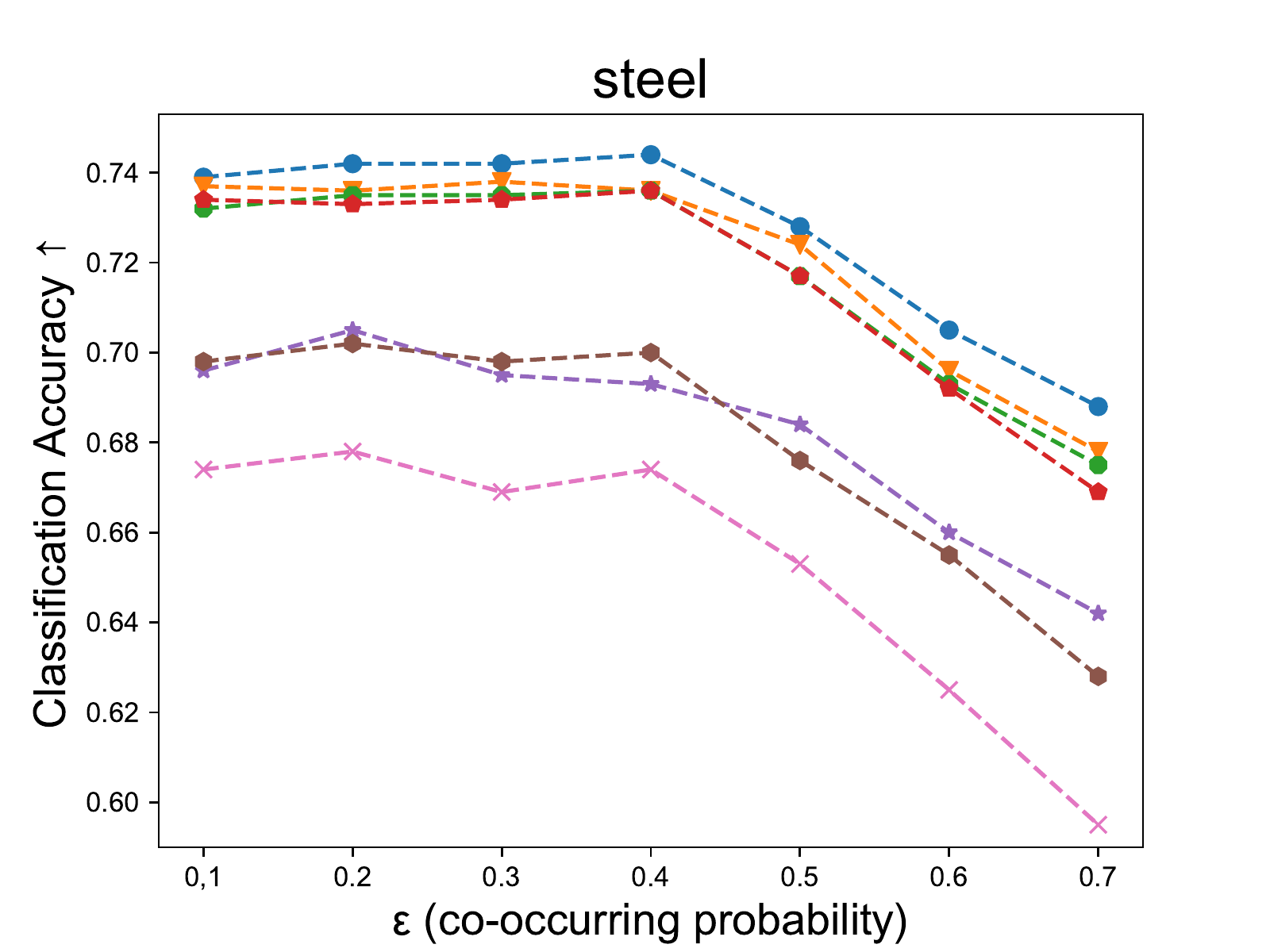}
    }\\
    \subfigure[Classification accuracy on controlled UCI data sets with $p$ varying from 0.1 to 0.7 ($r=1$).]{
	\includegraphics[scale=0.263]{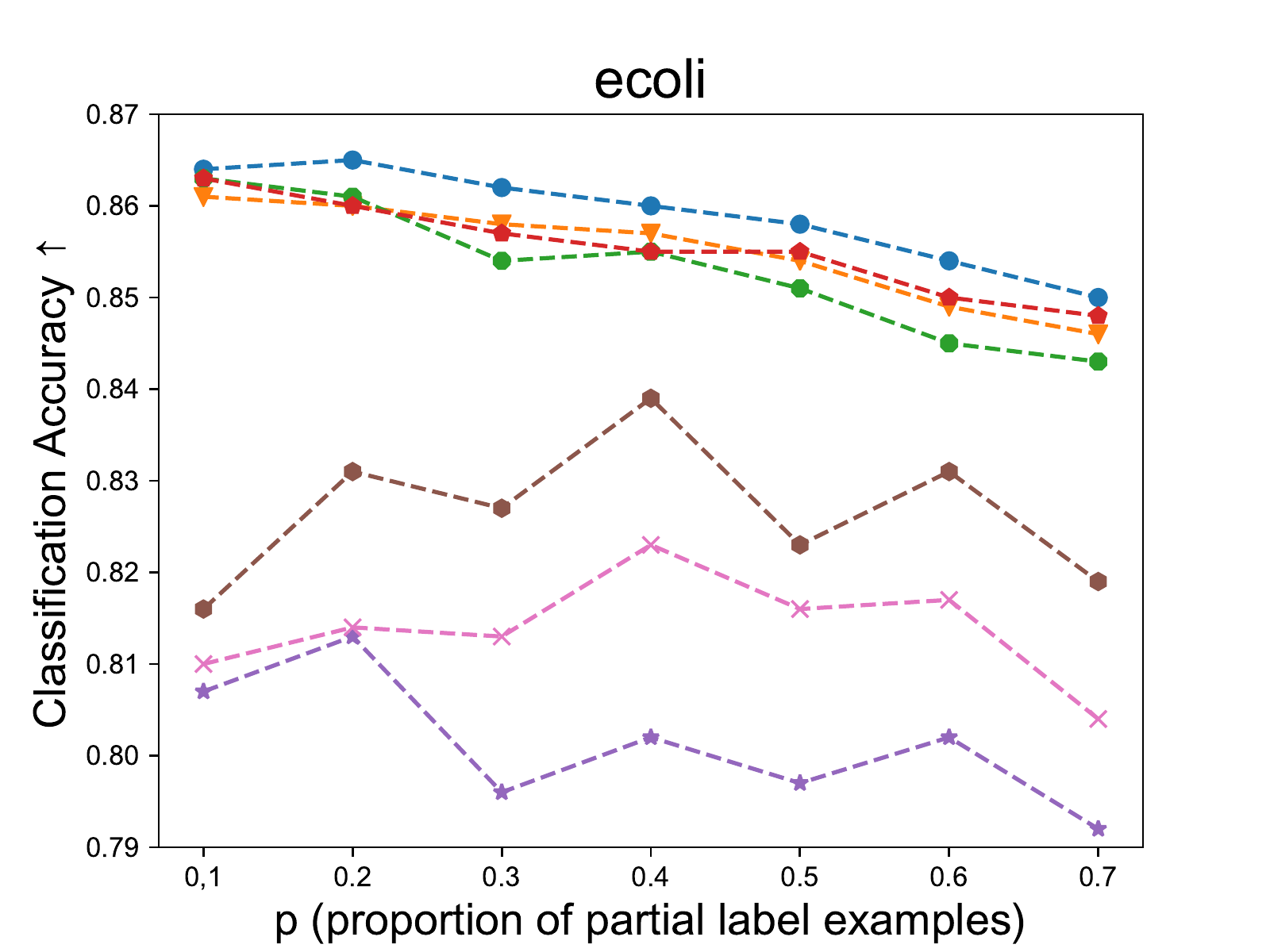}
	\includegraphics[scale=0.263]{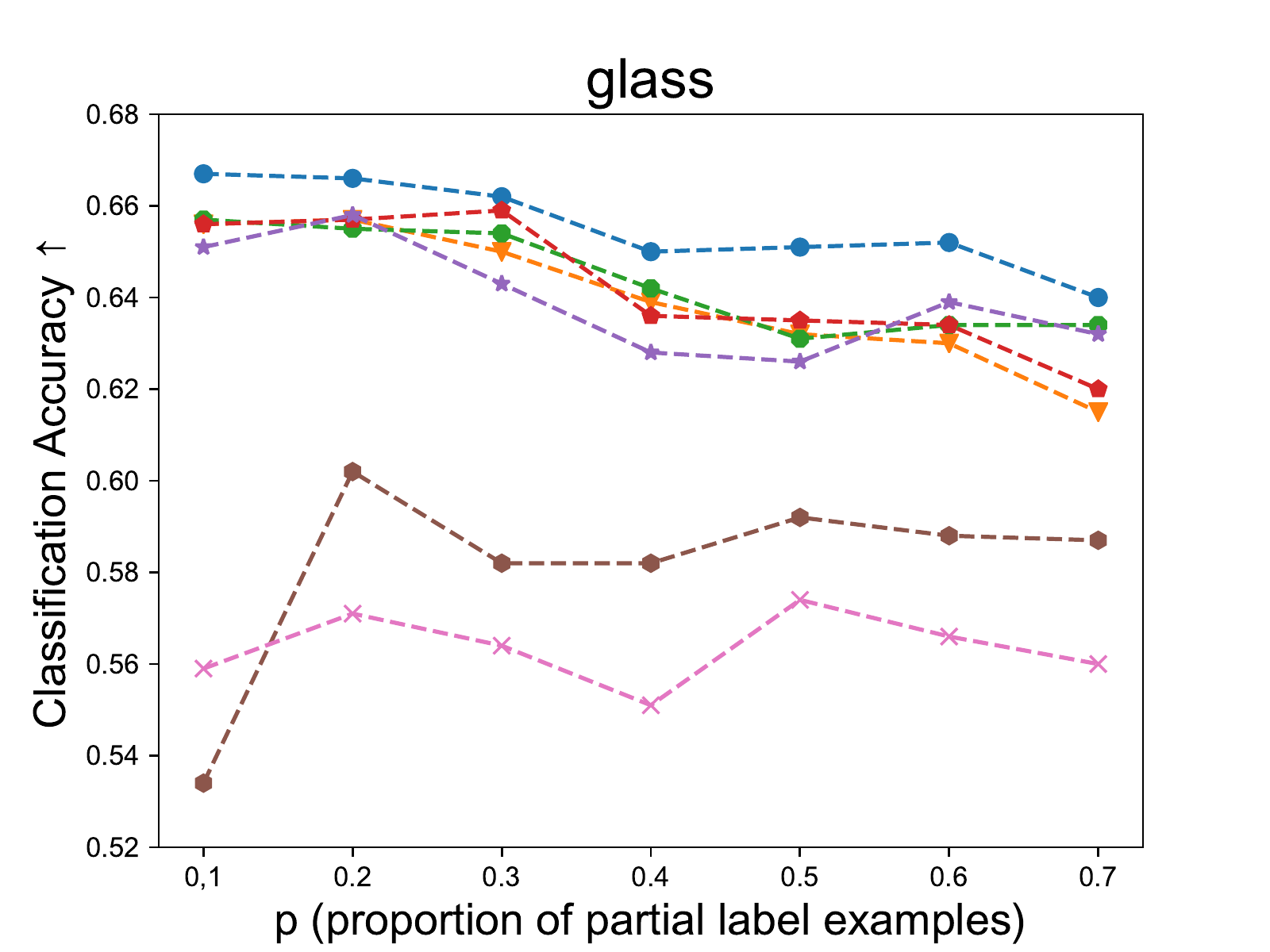}
	\includegraphics[scale=0.263]{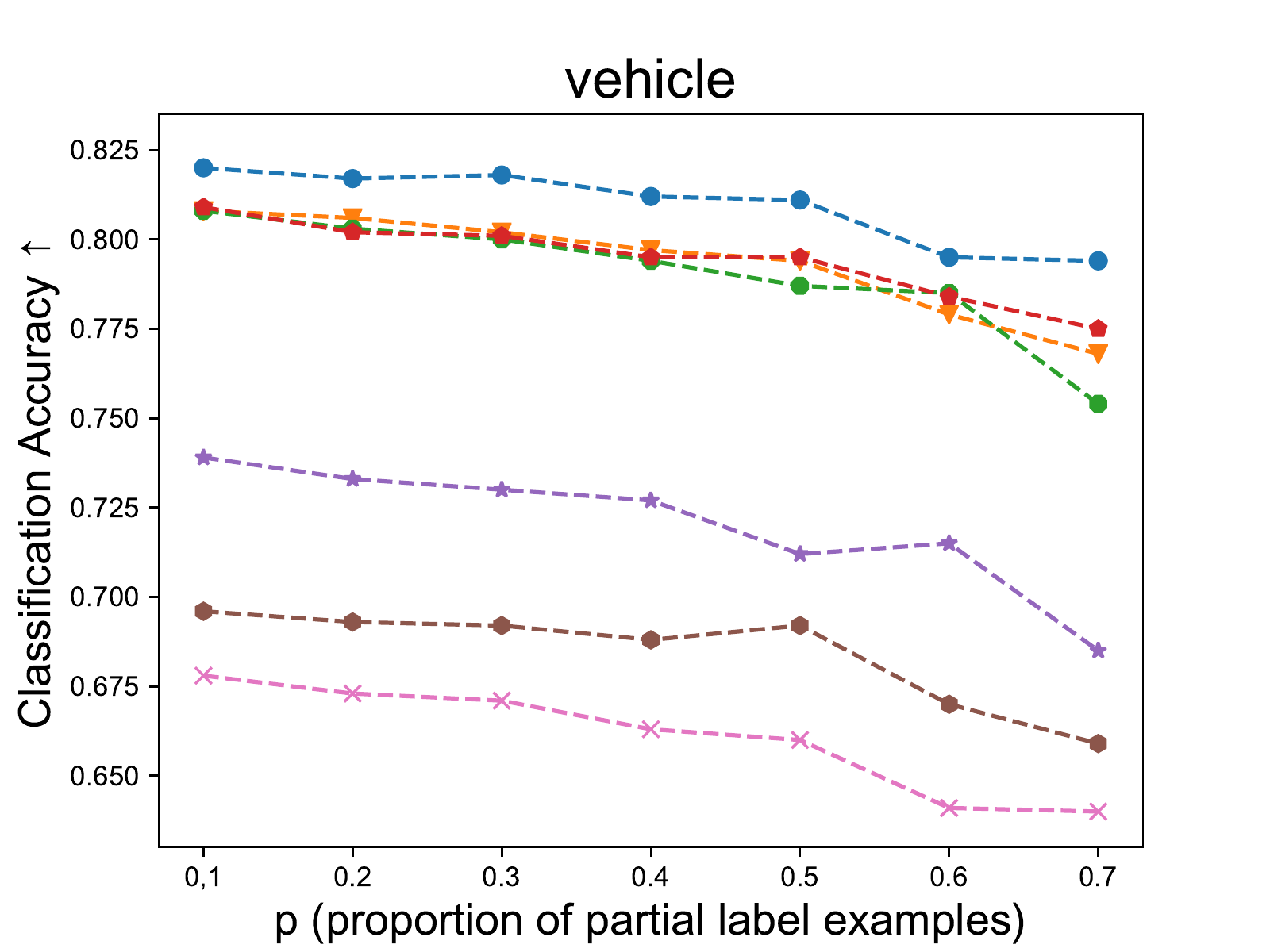}
	\includegraphics[scale=0.263]{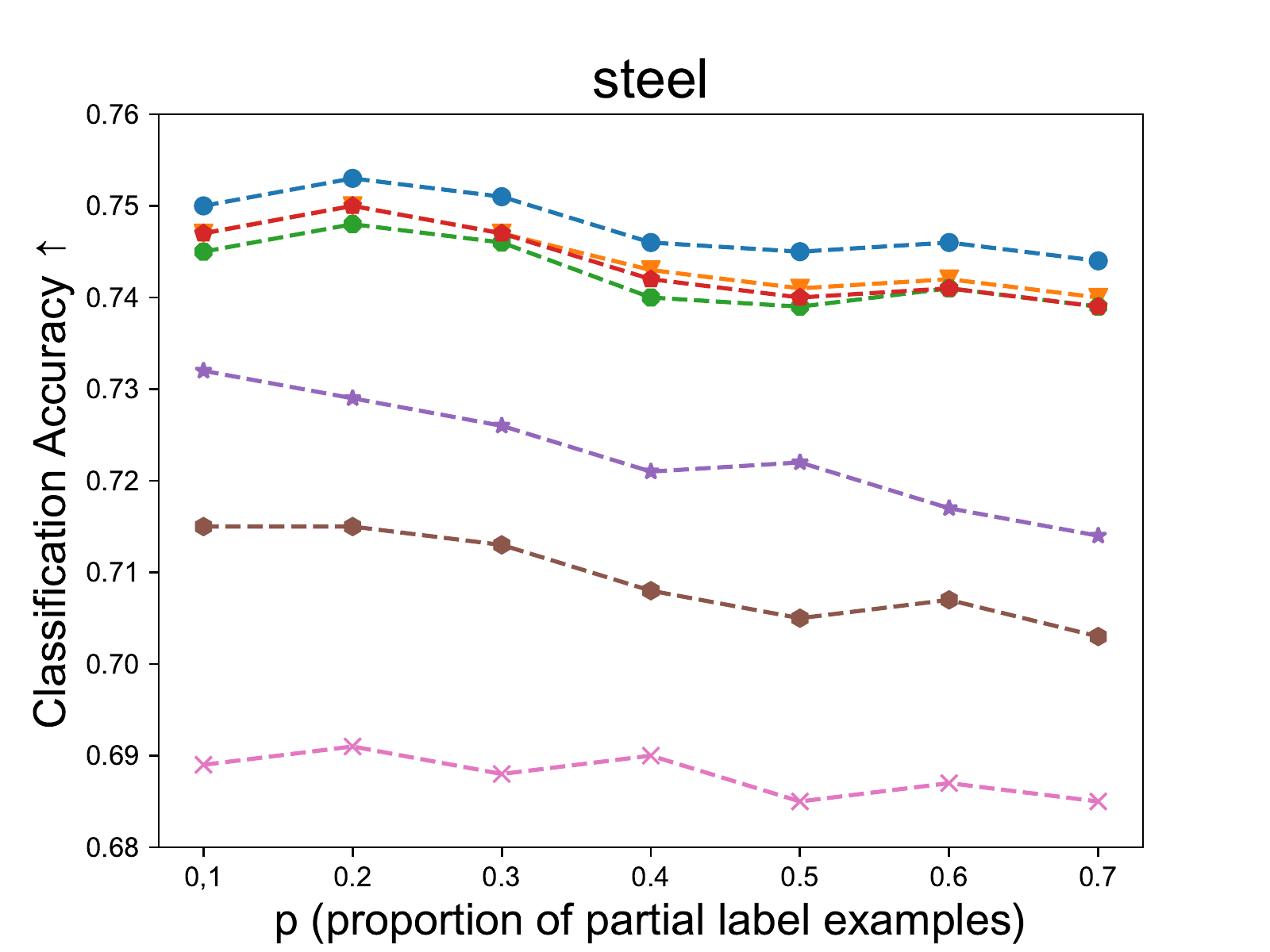}}\\
	\subfigure[Classification accuracy on controlled UCI data sets with $p$ varying from 0.1 to 0.7 ($r=2$).]{
	\includegraphics[scale=0.263]{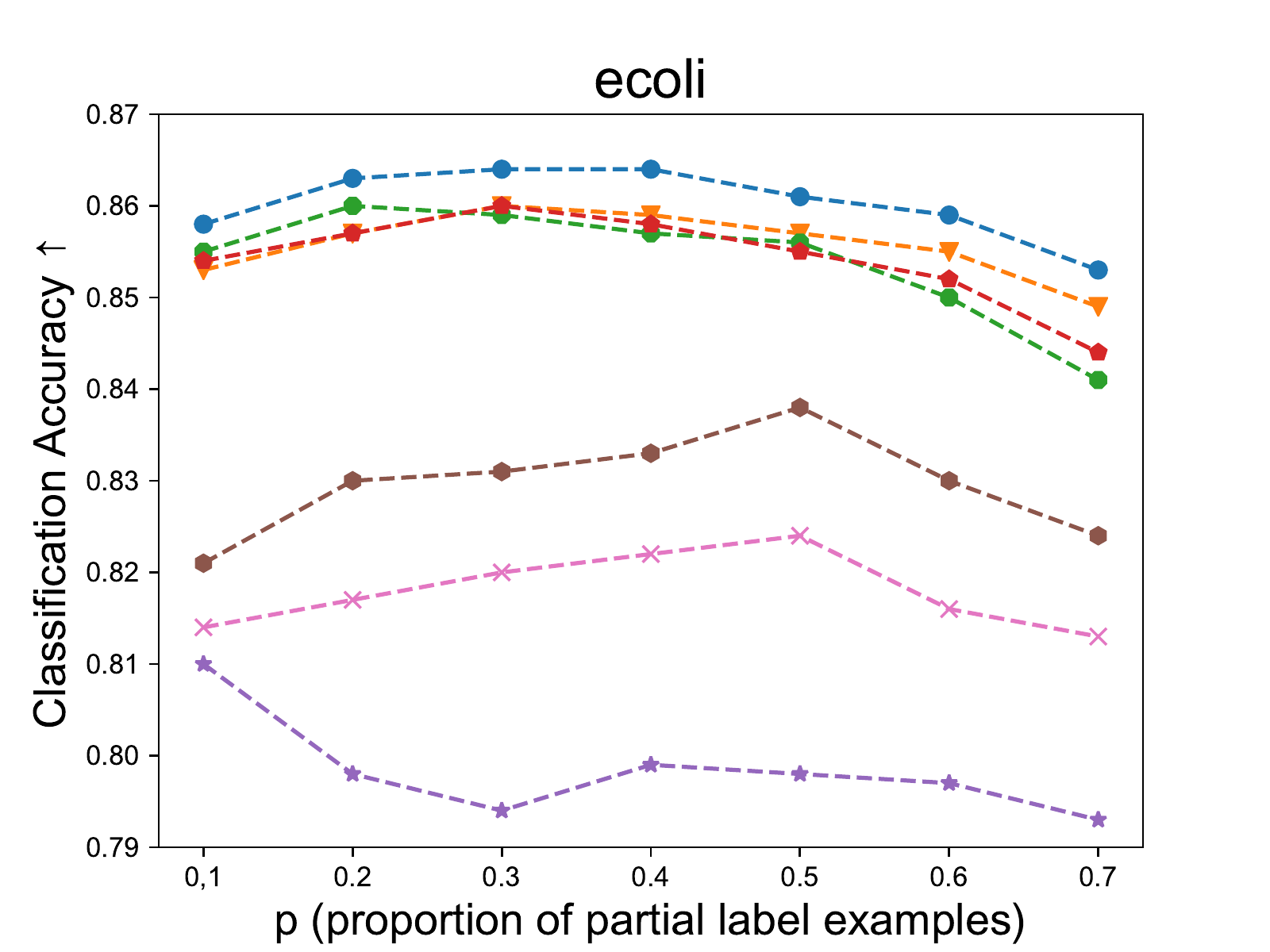}
	\includegraphics[scale=0.263]{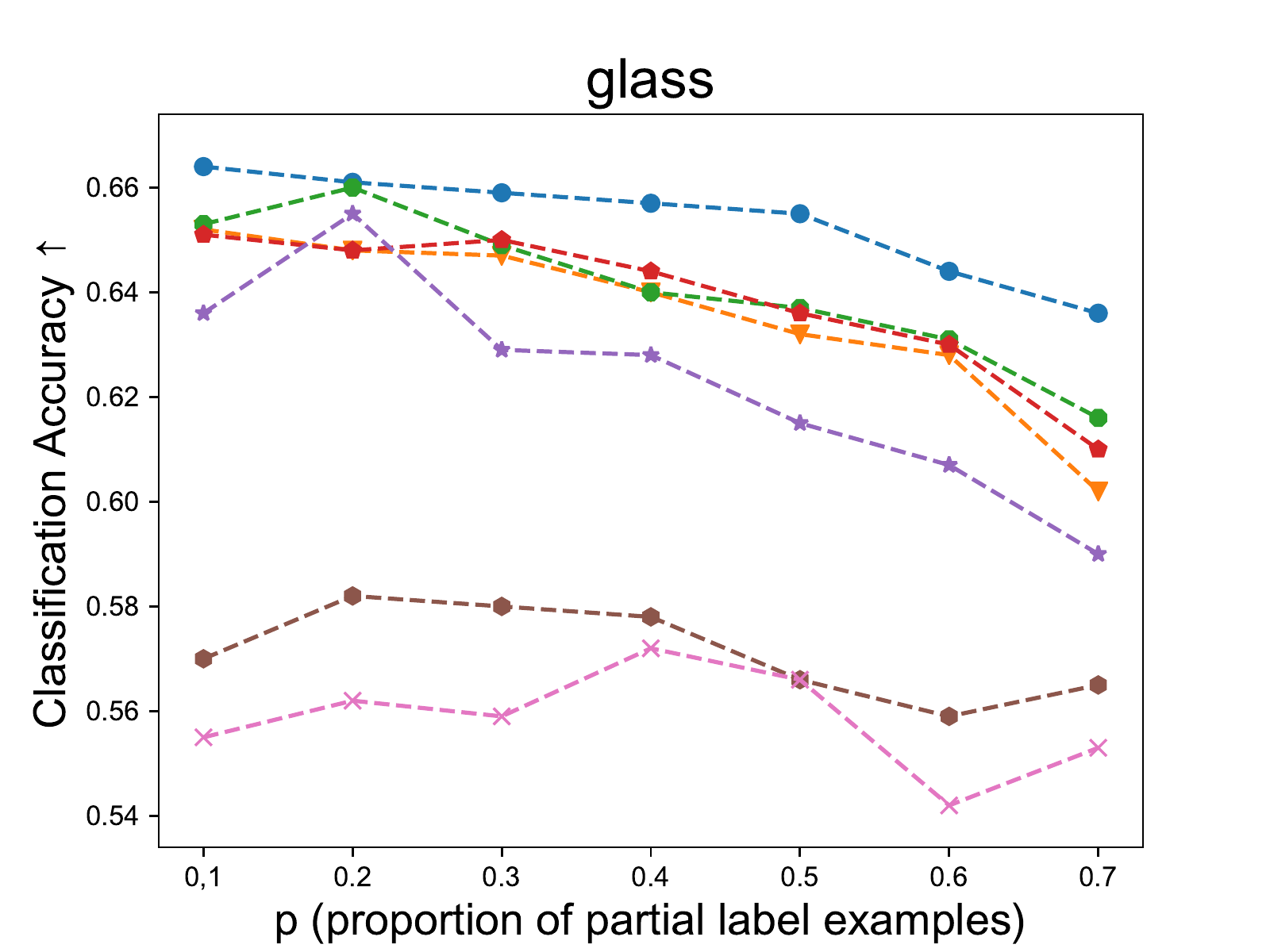}
	\includegraphics[scale=0.263]{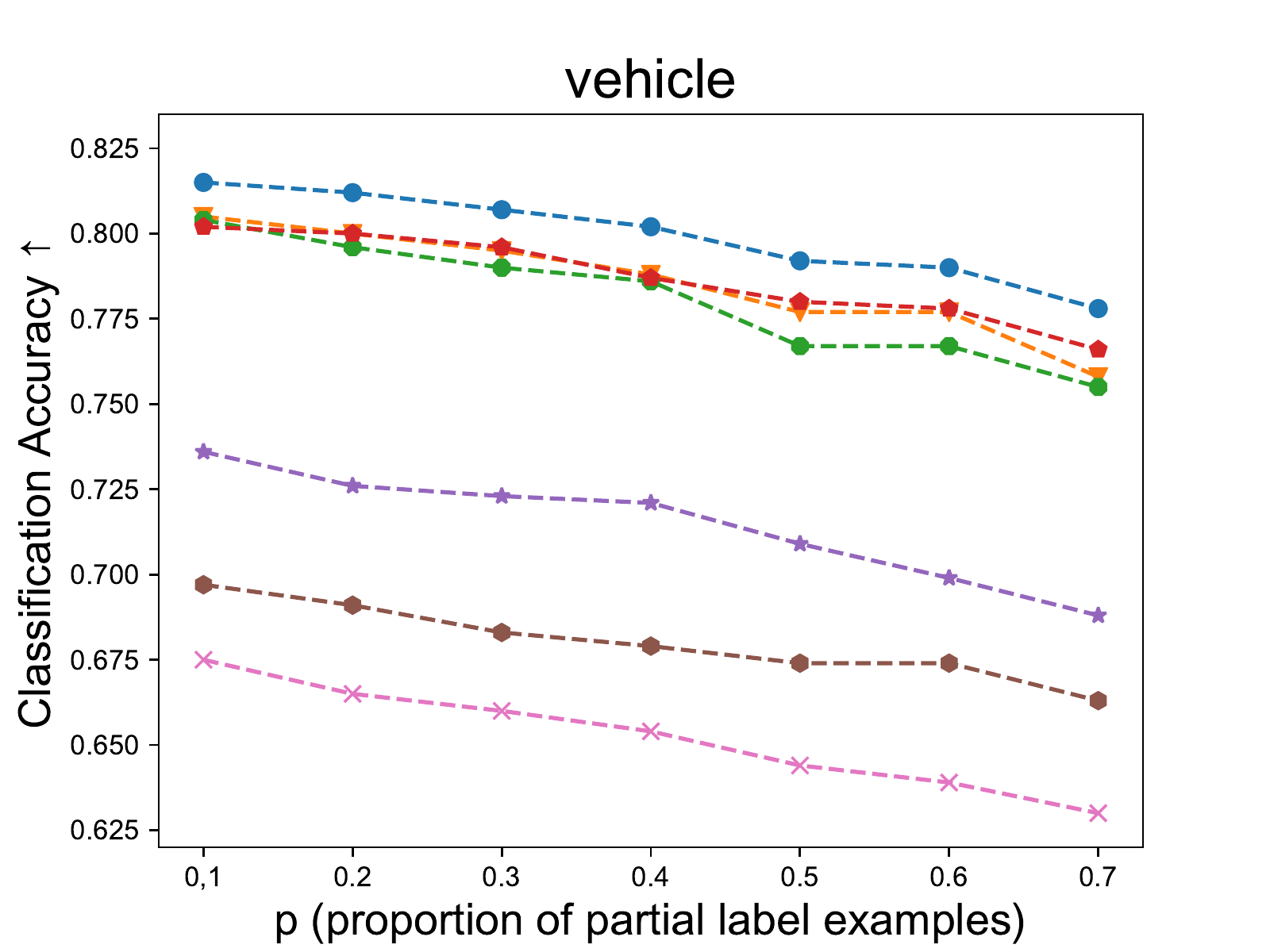}
	\includegraphics[scale=0.263]{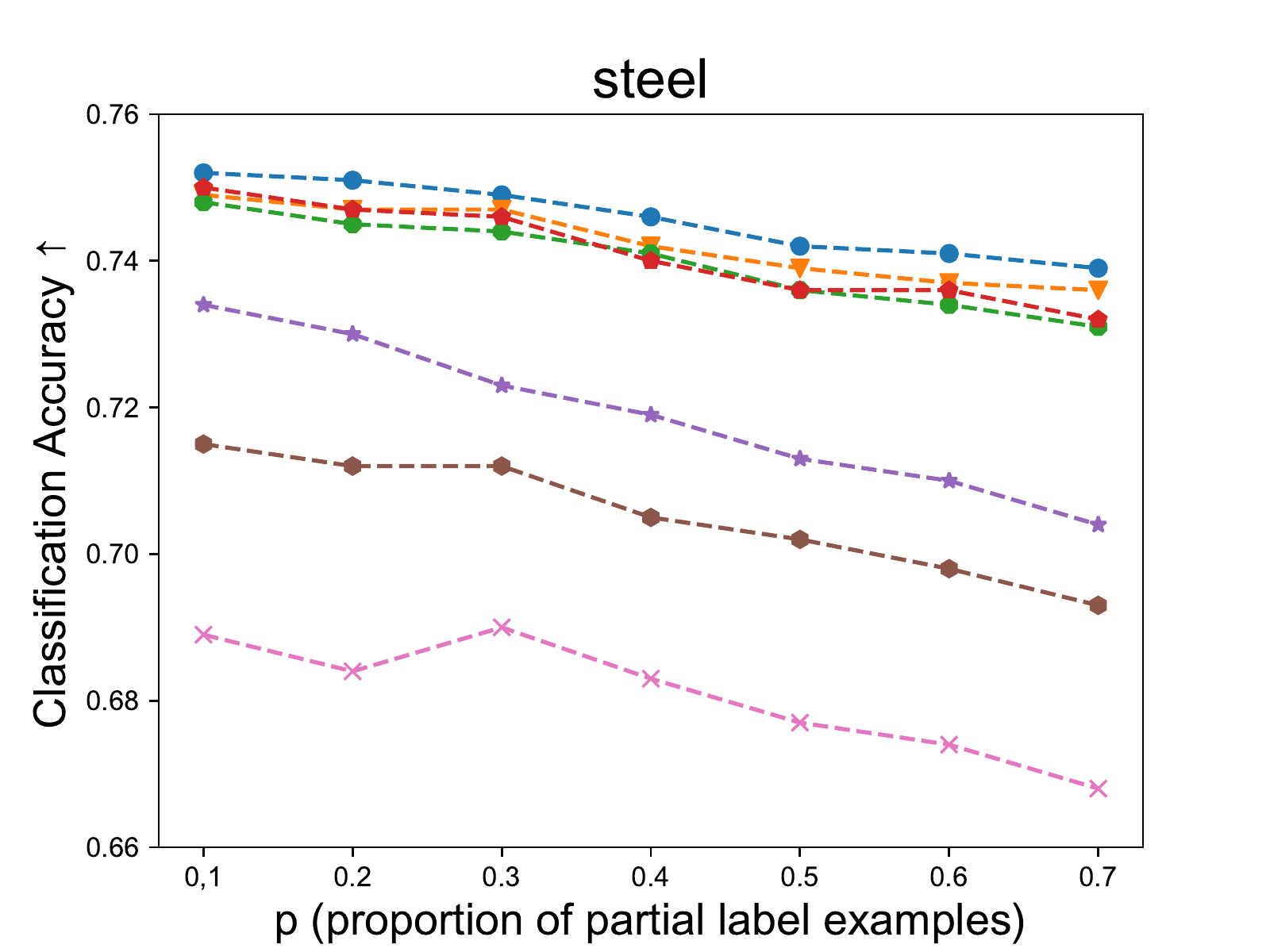}}\\
	\subfigure[Classification accuracy on controlled UCI data sets with $p$ varying from 0.1 to 0.7 ($r=3$).]{
	\includegraphics[scale=0.263]{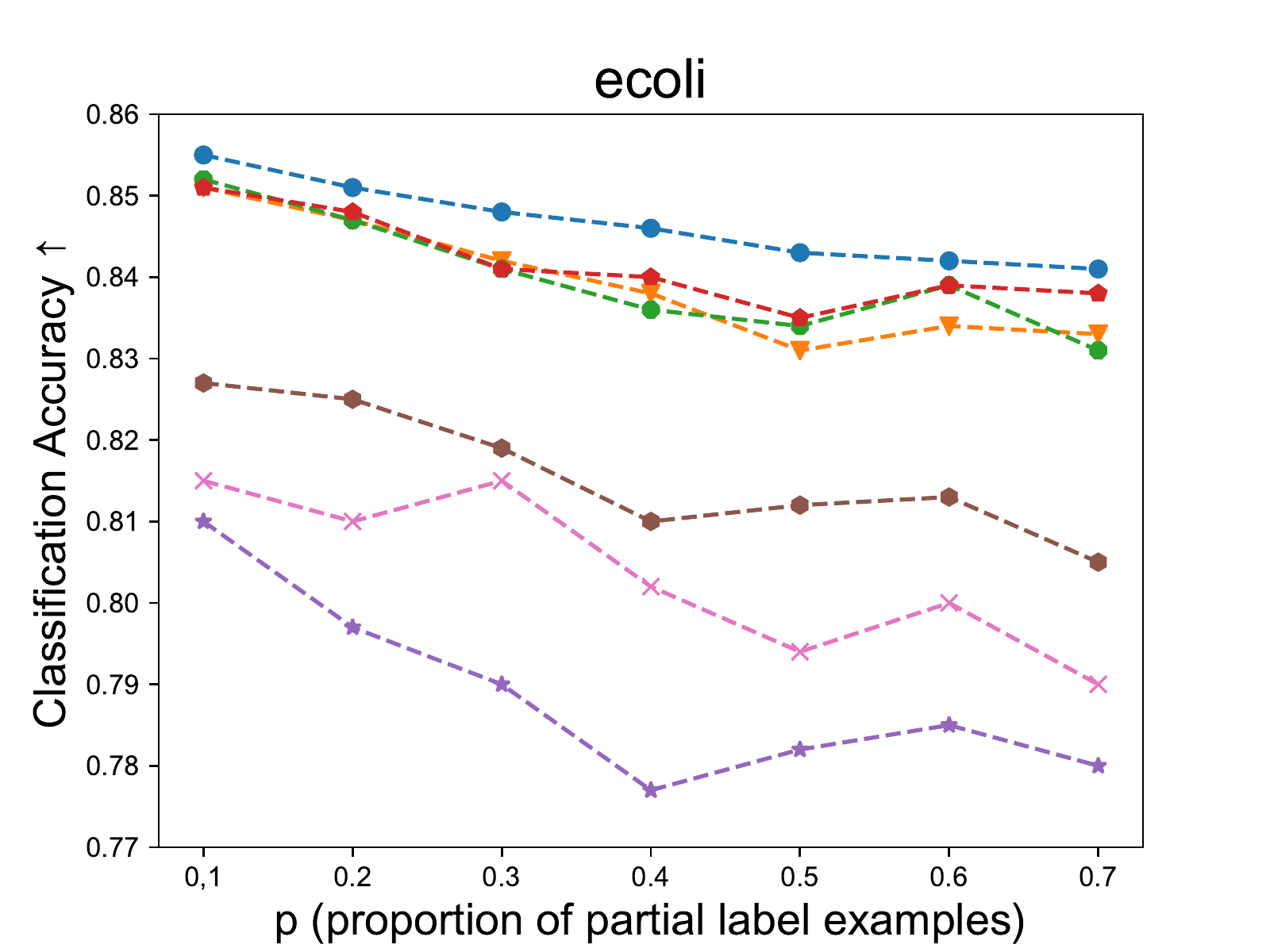}
	\includegraphics[scale=0.263]{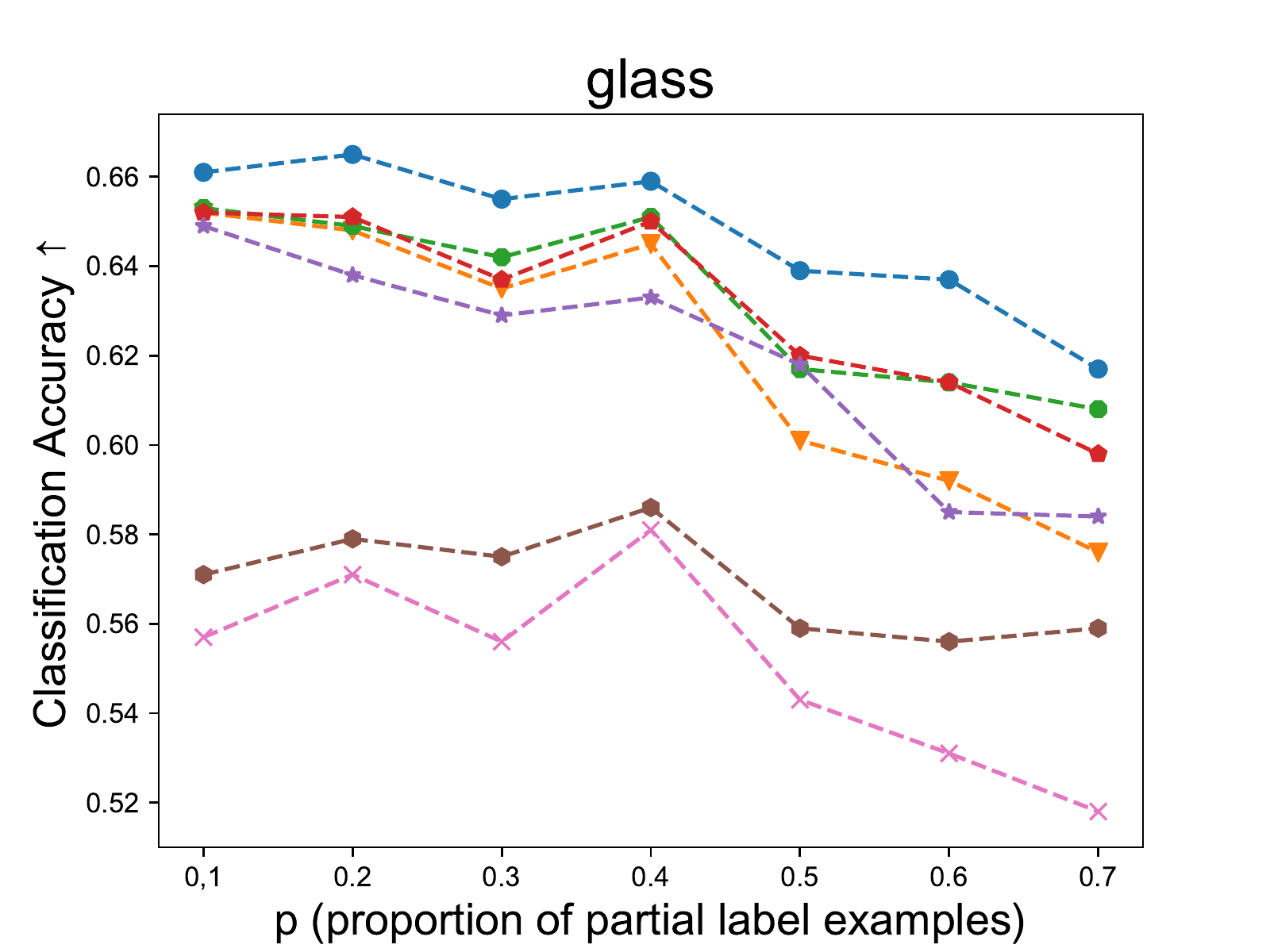}
	\includegraphics[scale=0.263]{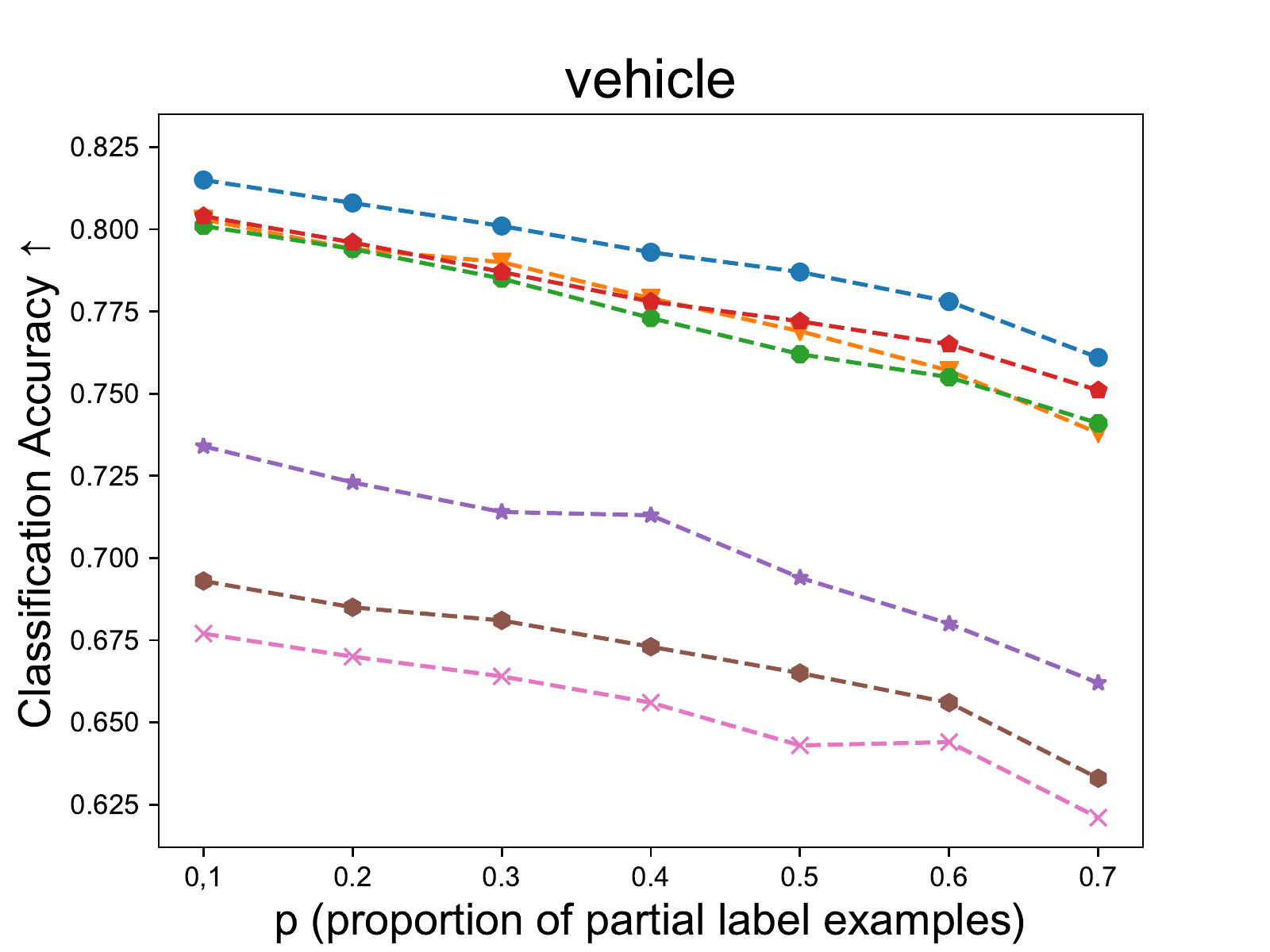}
	\includegraphics[scale=0.263]{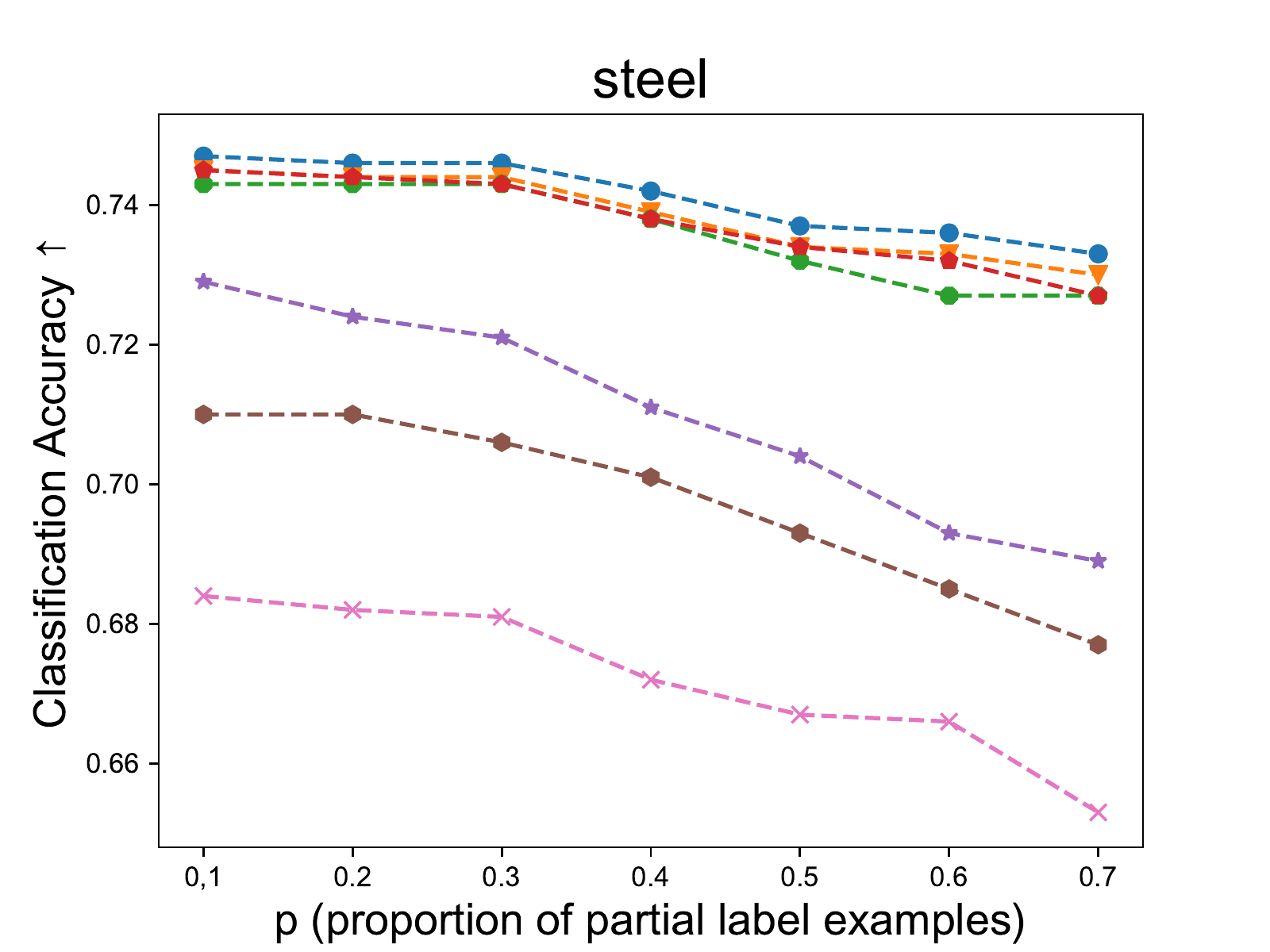}}
	\caption{Classification accuracy on the controlled UCI data sets.}
	\label{fig performance uci} 
\end{figure*}

\begin{table*}[ht!]\footnotesize

\setlength{\tabcolsep}{1.8mm}
\renewcommand\arraystretch{1} 
    \centering
    \caption{The comparison of different methods on the real-world data sets on transductive accuracy. $\bullet$/$\circ$ indicates whether PL-CL is statistically superior/inferior to the compared algorithm according to pairwise $t$-test at significance level of 0.05.}
    \begin{tabular}{|c l l l l l l l l|}\hline\hline
         Data set & \multicolumn{1}{c}{FG-NET} & \multicolumn{1}{c}{Lost} & \multicolumn{1}{c}{MSRCv2} & \multicolumn{1}{c}{Mirflickr} & \multicolumn{1}{c}{Soccer Player} &
         \multicolumn{1}{c}{Yahoo!News}&
         \multicolumn{1}{c}{FG-NET(MAE3)}&
         \multicolumn{1}{c|}{FG-NET(MAE5)}\\\hline
         
         PL-CL & 0.159 $\pm$ 0.016 & 0.832 $\pm$ 0.019  & 0.585 $\pm$ 0.012 & 0.697 $\pm$ 0.016 & 0.715 $\pm$ 0.002 & 0.840 $\pm$ 0.004 & 0.600 $\pm$ 0.023 & 0.737 $\pm$ 0.018 \\
         SURE & 0.158 $\pm$ 0.012 & 0.798 $\pm$ 0.019 $\bullet$ &  0.603 $\pm$ 0.015 $\circ$ & 0.652 $\pm$ 0.022 $\bullet$ & 0.700 $\pm$ 0.003 $\bullet$ & 0.798 $\pm$ 0.005 $\bullet$ & 0.590 $\pm$ 0.018 $\bullet$ & 0.727 $\pm$ 0.019 $\bullet$ \\
         PL-AGGD & 0.130 $\pm$ 0.015 $\bullet$ & 0.804 $\pm$ 0.016 $\bullet$ & 0.551 $\pm$ 0.015 $\bullet$ & 0.653 $\pm$ 0.015 $\bullet$ & 0.698 $\pm$ 0.003 $\bullet$ & 0.817 $\pm$ 0.005 $\bullet$ & 0.530 $\pm$ 0.021 $\bullet$ & 0.679 $\pm$ 0.024 $\bullet$ \\ 
         LALO & 0.153 $\pm$ 0.016 $\bullet$ & 0.817 $\pm$ 0.012 $\bullet$ & 0.548 $\pm$ 0.009 $\bullet$ & 0.675 $\pm$ 0.017 $\bullet$ & 0.698 $\pm$ 0.003 $\bullet$ & 0.821 $\pm$ 0.004 $\bullet$ & 0.592 $\pm$ 0.024 $\bullet$ & 0.730 $\pm$ 0.015 $\bullet$ \\
         IPAL & 0.148 $\pm$ 0.021 & 0.793 $\pm$ 0.017 $\bullet$ & 0.680 $\pm$ 0.013 $\circ$ & 0.586 $\pm$ 0.007 $\bullet$ & 0.681 $\pm$ 0.003 $\bullet$ & 0.839 $\pm$ 0.003 &  0.563 $\pm$ 0.021 $\bullet$ & 0.698 $\pm$ 0.022 $\bullet$ \\
         PLDA & 0.042 $\pm$ 0.005 $\bullet$ & 0.351 $\pm$ 0.060 $\bullet$ & 0.479 $\pm$ 0.015 $\bullet$ & 0.564 $\pm$ 0.015 $\bullet$ & 0.493 $\pm$ 0.004 $\bullet$ & 0.460 $\pm$ 0.009 $\bullet$ & 0.150 $\pm$ 0.012 $\bullet$ & 0.232 $\pm$ 0.012 $\bullet$ \\
         PL-KNN & 0.041 $\pm$ 0.007 $\bullet$ & 0.338 $\pm$ 0.016 $ \bullet$ & 0.415 $\pm$ 0.013 $\bullet$ & 0.466 $\pm$ 0.013 $\bullet$ & 0.504 $\pm$ 0.005 $\bullet$ & 0.403 $\pm$ 0.009 $\bullet$ & 0.285 $\pm$ 0.016 $\bullet$ & 0.438 $\pm$ 0.014 $\bullet$ \\\hline\hline
    \end{tabular}
   
    \label{tab:real world statistic transductive}
\end{table*}

\begin{table*}[ht!]\scriptsize

\setlength{\tabcolsep}{1mm}
\renewcommand\arraystretch{1} 
    \centering
    \caption{Ablation study of the proposed approach on the real-world data sets on classification accuracy. $\bullet$/$\circ$ indicates whether PL-CL is statistically superior/inferior to the compared algorithm according to pairwise $t$-test at significance level of 0.05.}
    
    \begin{tabular}{|c c c| l l l l l l l l |c|}\hline\hline
     \multicolumn{12}{|c|}{Classification Accuracy}\\\hline
         Kernel & Complementary Classifier & Graph & \multicolumn{1}{c}{FG-NET} & \multicolumn{1}{c}{Lost} & \multicolumn{1}{c}{MSRCv2} & \multicolumn{1}{c}{Mirflickr} & \multicolumn{1}{c}{Soccer Player} &
         \multicolumn{1}{c}{Yahoo!News}&
         \multicolumn{1}{c}{FG-NET(MAE3)}&
         \multicolumn{1}{c|}{FG-NET(MAE5)} & Average\\\hline

         \XSolid &  \XSolid  & \XSolid & 0.061 $\pm$ 0.006 $\bullet$ & 0.622 $\pm$ 0.019 $\bullet$ & 0.381 $\pm$ 0.015 $\bullet$ &  0.249 $\pm$ 0.010 $\bullet$ & 0.492 $\pm$ 0.003 $\bullet$ & 0.430 $\pm$ 0.051 $\bullet$ & 0.402 $\pm$ 0.031 $\bullet$ & 0.551 $\pm$ 0.024 $\bullet$ & 0.398\\

         \Checkmark & \XSolid & \XSolid & 0.060 $\pm$ 0.008 $\bullet$ & 0.654 $\pm$ 0.019 $\bullet$ & 0.426 $\pm$ 0.017 $\bullet$ & 0.533 $\pm$ 0.012 $\bullet$ & 0.515 $\pm$ 0.010 $\bullet$ & 0.526 $\pm$ 0.006 $\bullet$ & 0.413 $\pm$ 0.026 $\bullet$ & 0.564 $\pm$ 0.018 $\bullet$ & 0.461\\

         \Checkmark & \XSolid & \Checkmark & 0.057 $\pm$ 0.009 $\bullet$ & 0.705 $\pm$ 0.023 $\bullet$ & 0.462 $\pm$ 0.015 $\bullet$ & 0.637 $\pm$ 0.011 $\bullet$ & 0.530 $\pm$ 0.004 $\bullet$ & 0.517 $\pm$ 0.046 $\bullet$ & 0.416 $\pm$ 0.017 $\bullet$ & 0.560 $\pm$ 0.019 $\bullet$ & 0.485 \\
         \Checkmark & \Checkmark & \XSolid  & 0.065 $\pm$ 0.008 $\bullet$ & 0.684 $\pm$ 0.022 $\bullet$ & 0.456 $\pm$ 0.014 $\bullet$ & 0.635 $\pm$ 0.012 $\bullet$ & 0.529 $\pm$ 0.003 $\bullet$  & 0.607 $\pm$ 0.003 $\bullet$ & 0.426 $\pm$ 0.024 $\bullet$ & 0.566 $\pm$ 0.017 $\bullet$  & 0.496\\
         
          \Checkmark & \Checkmark &  \Checkmark & 0.072 $\pm$ 0.009 & 0.709 $\pm$ 0.022  & 0.469 $\pm$ 0.016 & 0.642 $\pm$ 0.012 & 0.534 $\pm$ 0.004 & 0.618 $\pm$ 0.003 & 0.433 $\pm$ 0.022 & 0.575 $\pm$ 0.015 & 0.507 \\
         \hline

          \multicolumn{12}{|c|}{Transductive Accuracy}\\\hline
         Kernel & Complementary Classifier & Graph  & \multicolumn{1}{c}{FG-NET} & \multicolumn{1}{c}{Lost} & \multicolumn{1}{c}{MSRCv2} & \multicolumn{1}{c}{Mirflickr} & \multicolumn{1}{c}{Soccer Player} &
         \multicolumn{1}{c}{Yahoo!News}&
         \multicolumn{1}{c}{FG-NET(MAE3)}&
         \multicolumn{1}{c|}{FG-NET(MAE5)} & Average\\\hline

         \XSolid & \XSolid & \XSolid & 0.161 $\pm$ 0.017 $\bullet$ & 0.729 $\pm$ 0.017 $\bullet$ & 0.414 $\pm$ 0.013 $\bullet$ & 0.464 $\pm$ 0.008 $\bullet$ & 0.492 $\pm$ 0.004 $\bullet$ & 0.439 $\pm$ 0.003 $\bullet$ & 0.580 $\pm$ 0.024 $\bullet$ & 0.717 $\pm$ 0.022 $\bullet$ & 0.499\\

          \Checkmark & \XSolid & \XSolid & 0.159 $\pm$ 0.012 $\bullet$ & 0.748 $\pm$ 0.015 $\bullet$ & 0.515 $\pm$ 0.018 $\bullet$ & 0.594 $\pm$ 0.013 $\bullet$ & 0.700 $\pm$ 0.004 $\bullet$ & 0.774 $\pm$ 0.032 $\bullet$ & 0.588 $\pm$ 0.021 $\bullet$ & 0.726 $\pm$ 0.010 $\bullet$ & 0.600\\
         
          \Checkmark & \XSolid &  \Checkmark  & 0.141 $\pm$ 0.011 $\bullet$ & 0.819 $\pm$ 0.017 $\bullet$ & 0.572 $\pm$ 0.013 $\bullet$ & 0.685 $\pm$ 0.014 $\bullet$ & 0.714 $\pm$ 0.002 $\bullet$ & 0.839 $\pm$ 0.004 $\bullet$ & 0.568 $\pm$ 0.025 $\bullet$ & 0.706 $\pm$ 0.022 $\bullet$ & 0.630 \\
          \Checkmark & \Checkmark & \XSolid  & 0.150 $\pm$ 0.015 $\bullet$ & 0.824 $\pm$ 0.021 $\bullet$ & 0.567 $\pm$ 0.008 $\bullet$ & 0.687 $\pm$ 0.012 $\bullet$ & 0.701 $\pm$ 0.003 $\bullet$  & 0.829 $\pm$ 0.004 $\bullet$ & 0.570 $\pm$ 0.022 $\bullet$ & 0.710 $\pm$ 0.021 $\bullet$ & 0.629\\
          \Checkmark & \Checkmark &  \Checkmark  & 0.159 $\pm$ 0.016 & 0.832 $\pm$ 0.019 & 0.585 $\pm$ 0.012 & 0.697 $\pm$ 0.016 & 0.715 $\pm$ 0.002 & 0.840 $\pm$ 0.004 & 0.600 $\pm$ 0.023 & 0.737 $\pm$ 0.018 & 0.646\\
          \hline\hline
    \end{tabular}
       
    \label{tab:real world further}
\end{table*}

\begin{figure*} [ht!]
\centering
	\subfigure[\label{fig:af}][$\alpha$]{
		\includegraphics[scale=0.208]{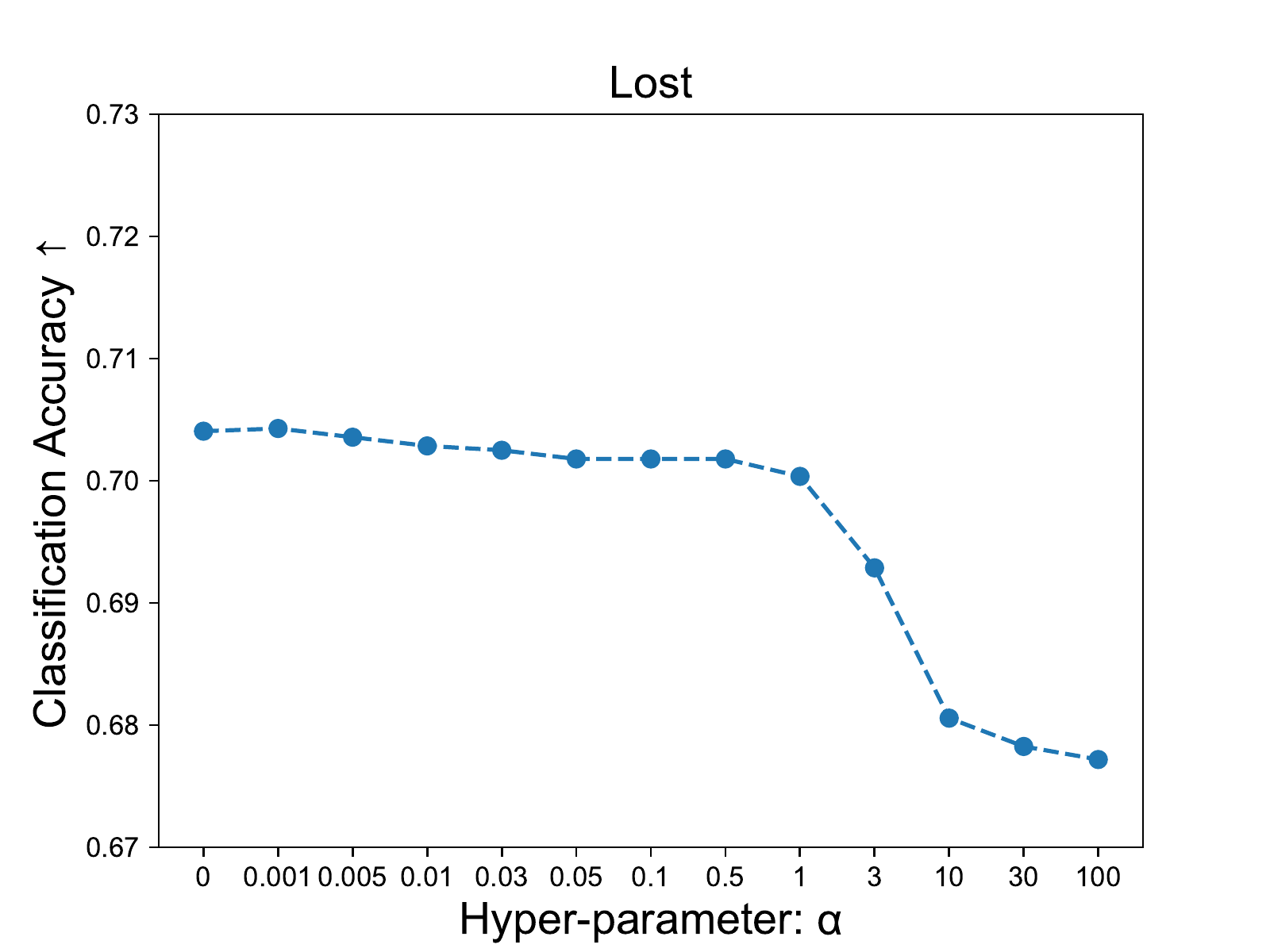}}
	\subfigure[\label{fig:bf}][$\beta$]{
		\includegraphics[scale=0.208]{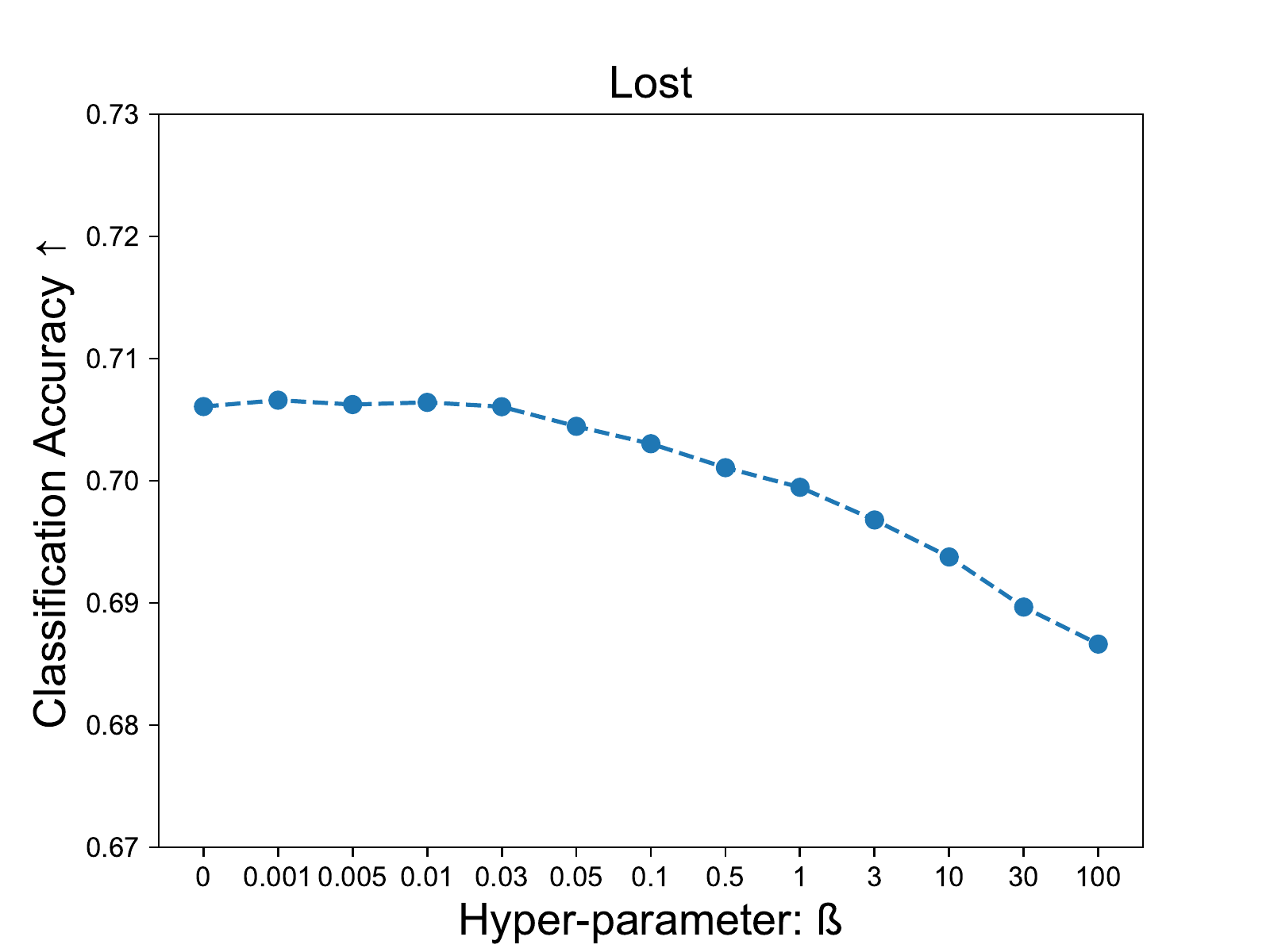}}
	\subfigure[\label{fig:cf}][$\gamma$]{
		\includegraphics[scale=0.208]{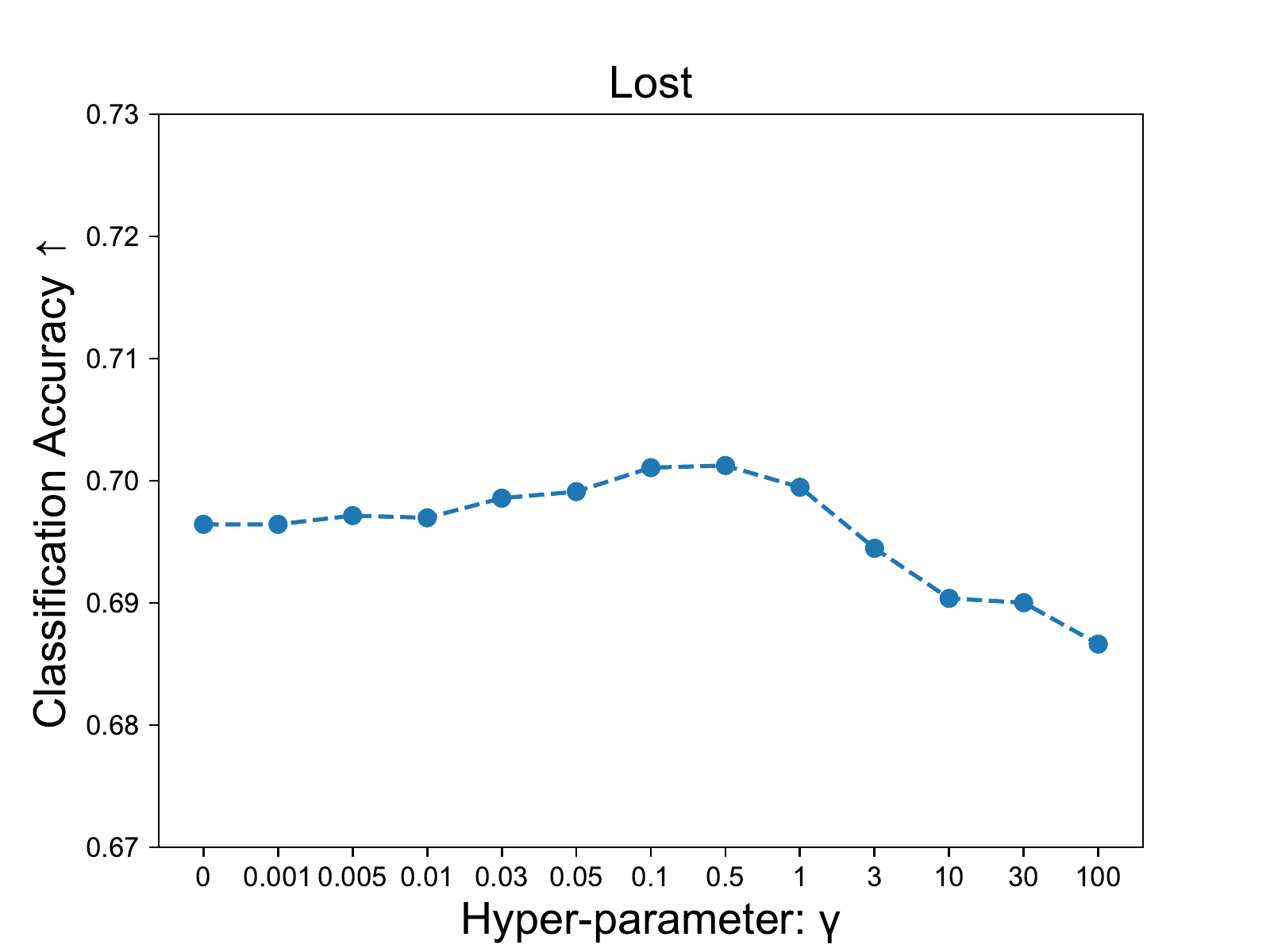}}
	\subfigure[\label{fig:gf}][$\mu$]{
		\includegraphics[scale=0.208]{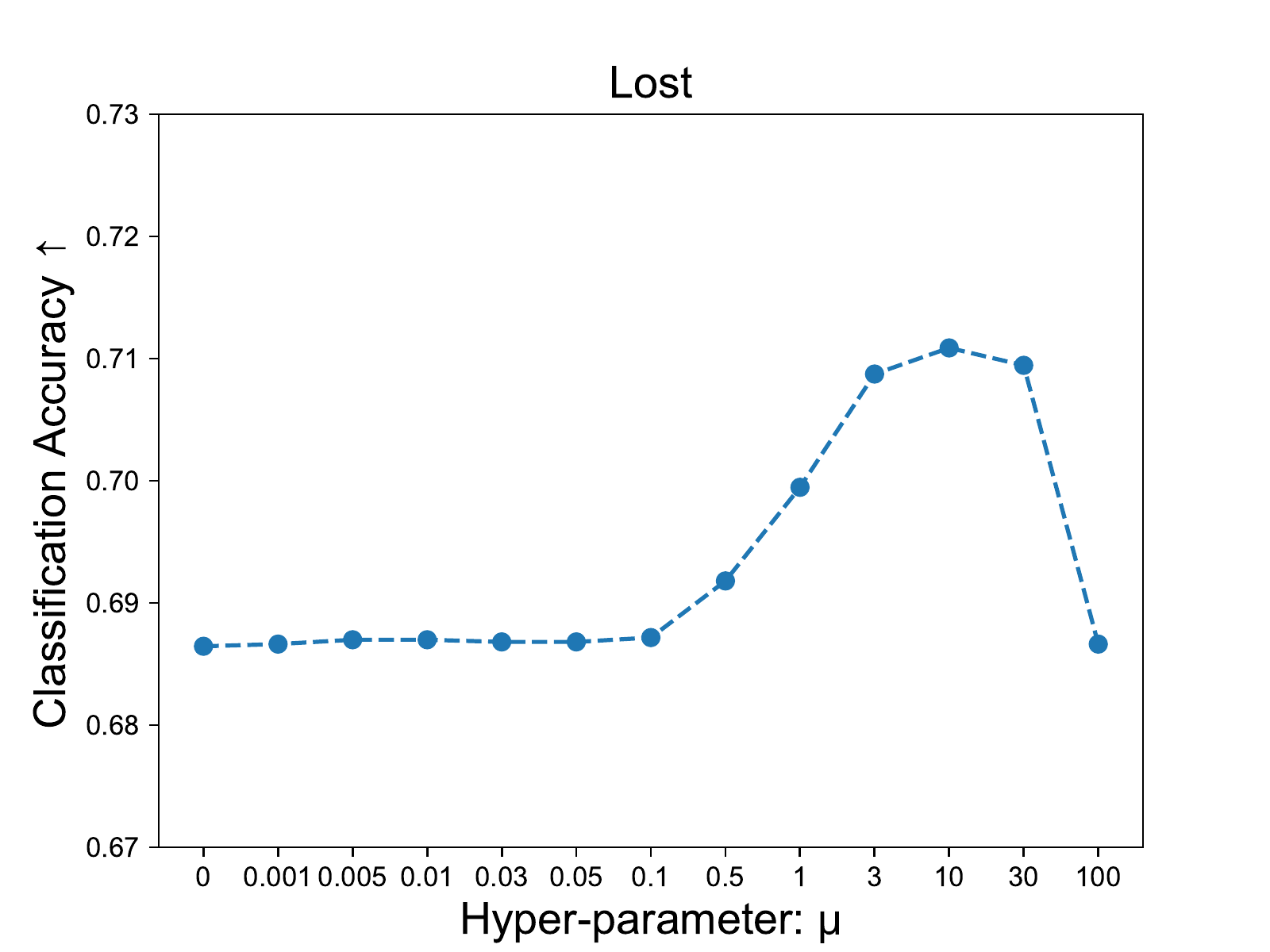}} 
    \subfigure[\label{fig:ff}][$\lambda$]{
		\includegraphics[scale=0.208]{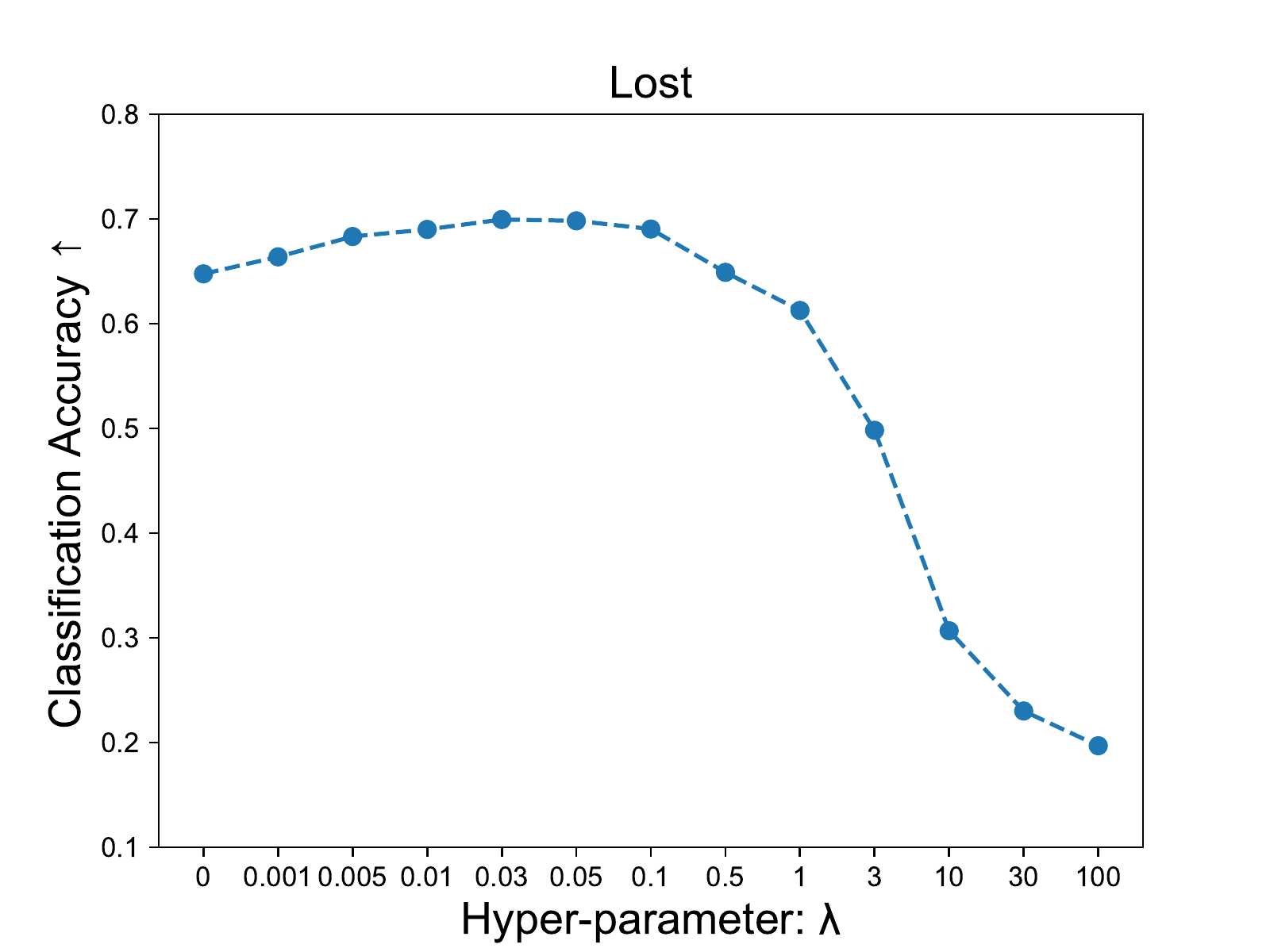}}
		\\
	\subfigure[\label{fig:gf}][FG-NET]{
		\includegraphics[scale=0.168]{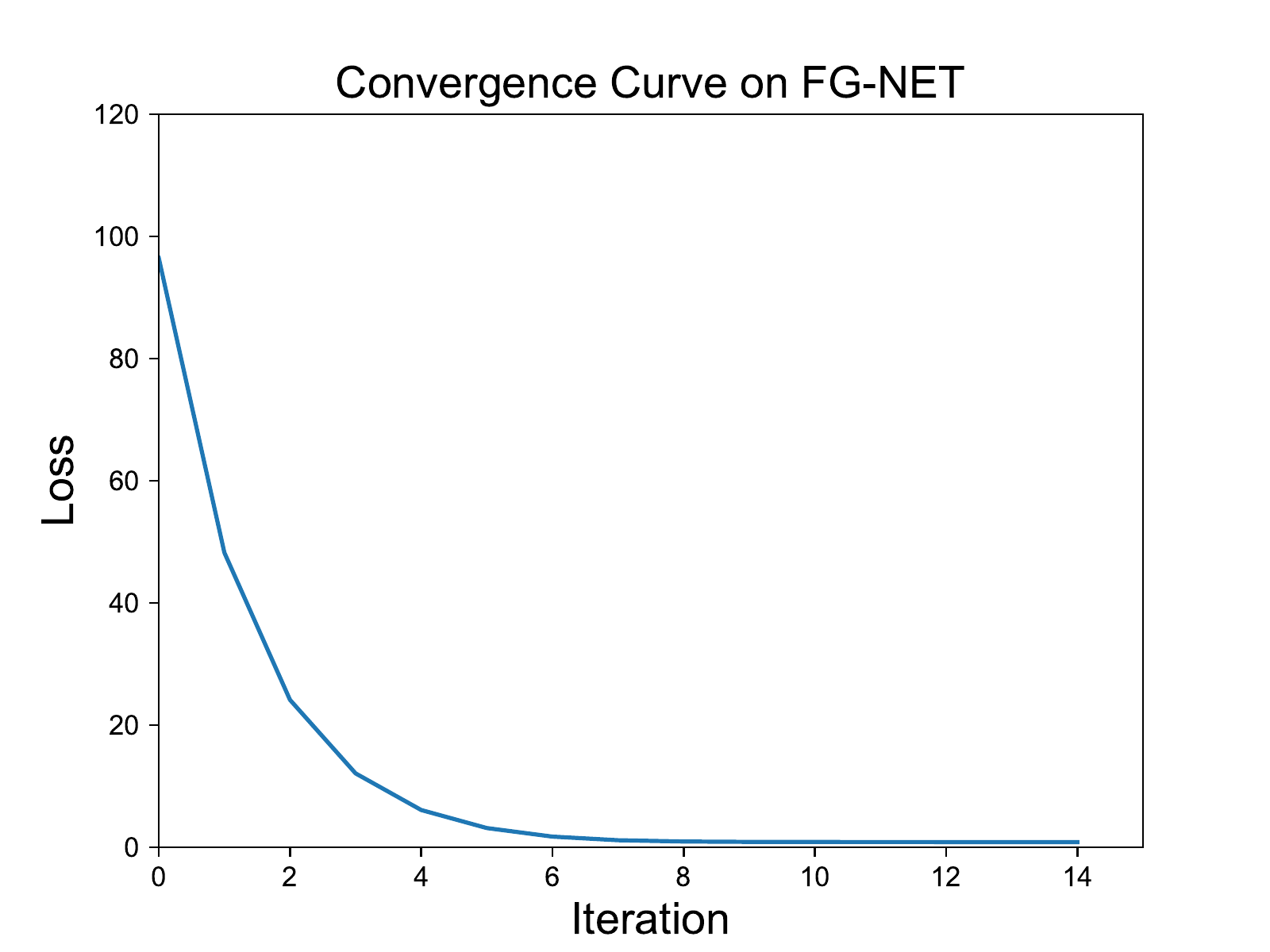} }
	\subfigure[\label{fig:gf}][Lost]{
		\includegraphics[scale=0.168]{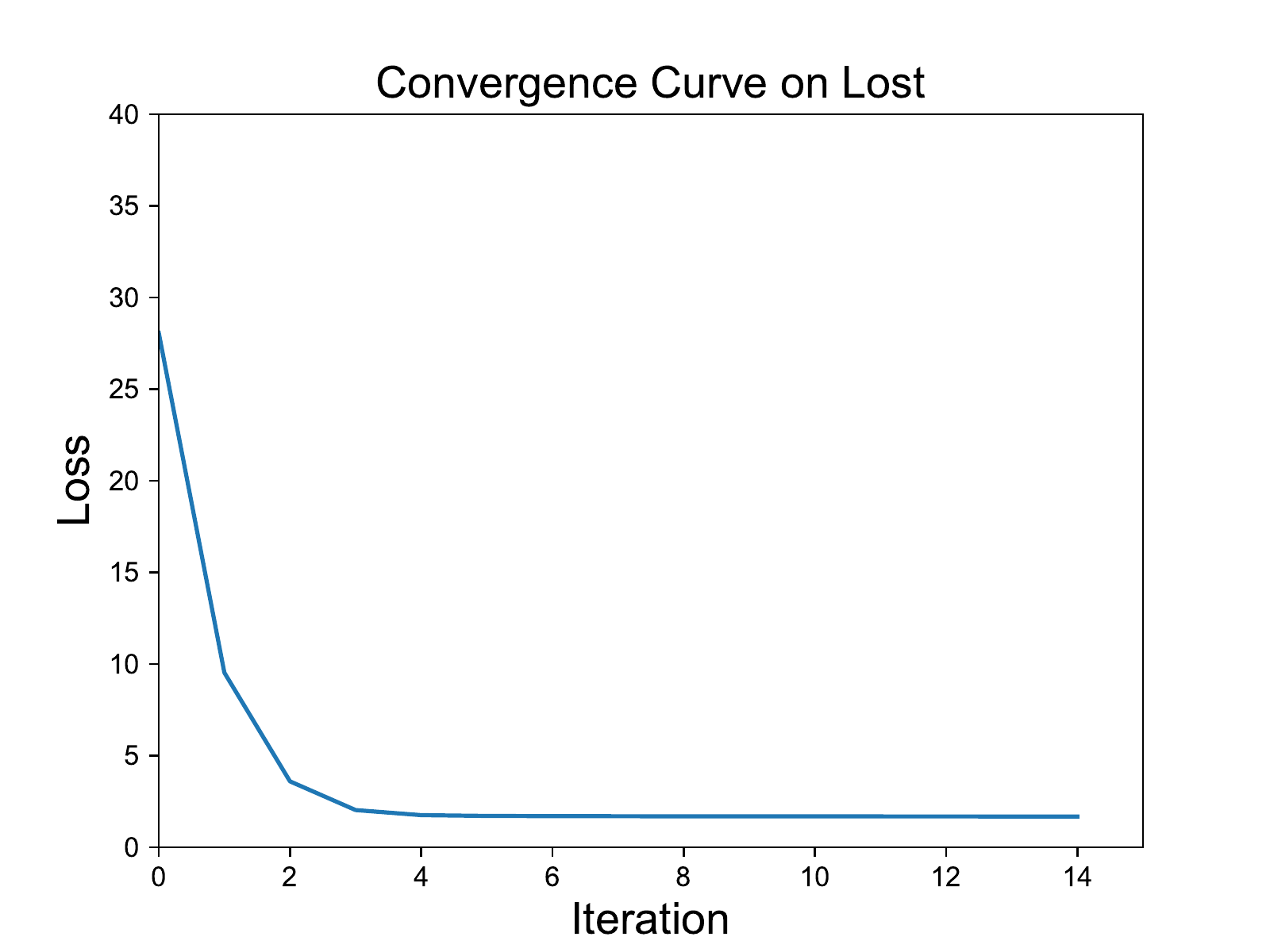} }
	\subfigure[\label{fig:gf}][MSRCv2]{
		\includegraphics[scale=0.168]{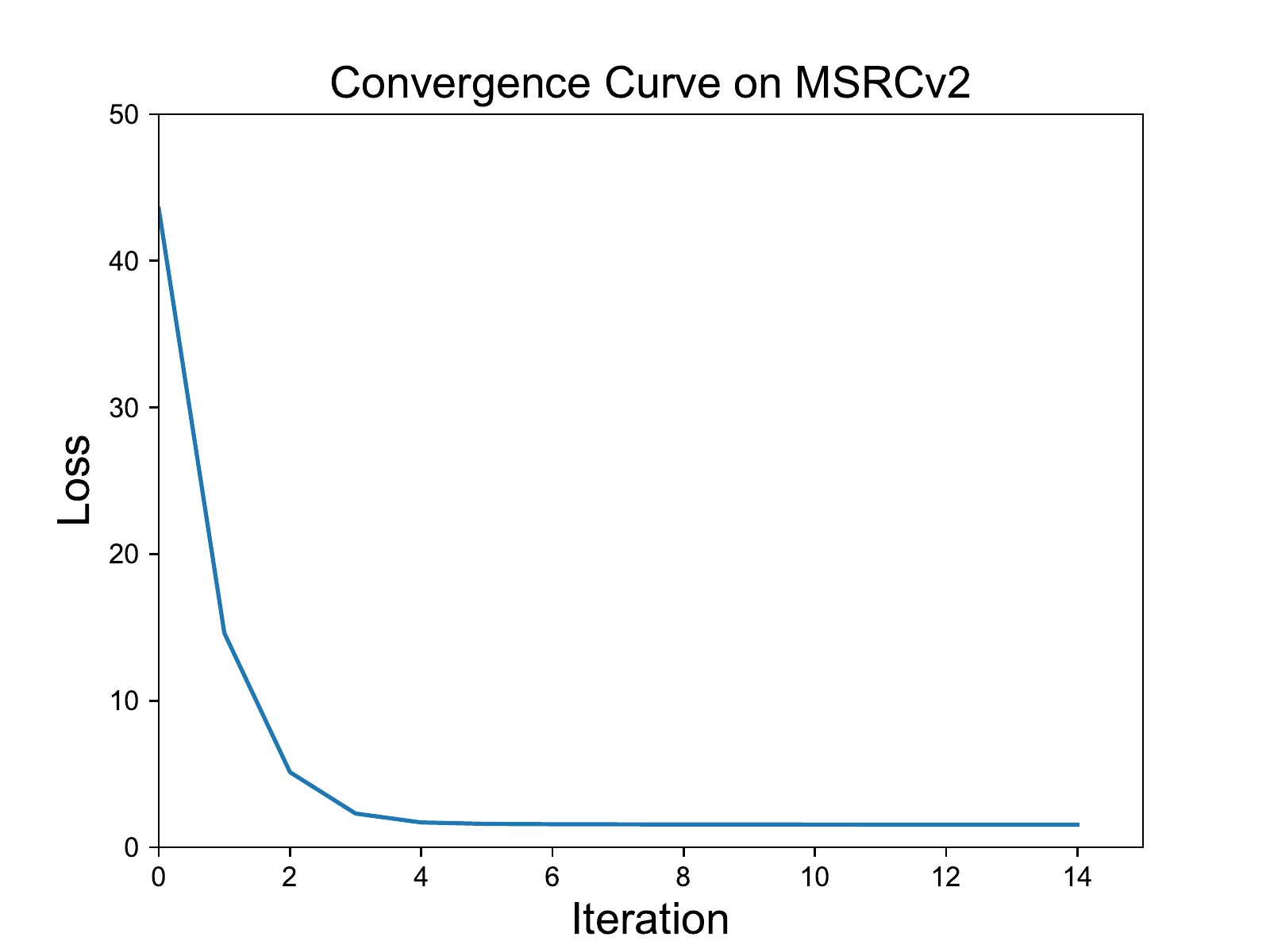}}
	\subfigure[\label{fig:gf}][Mirflickr]{
		\includegraphics[scale=0.168]{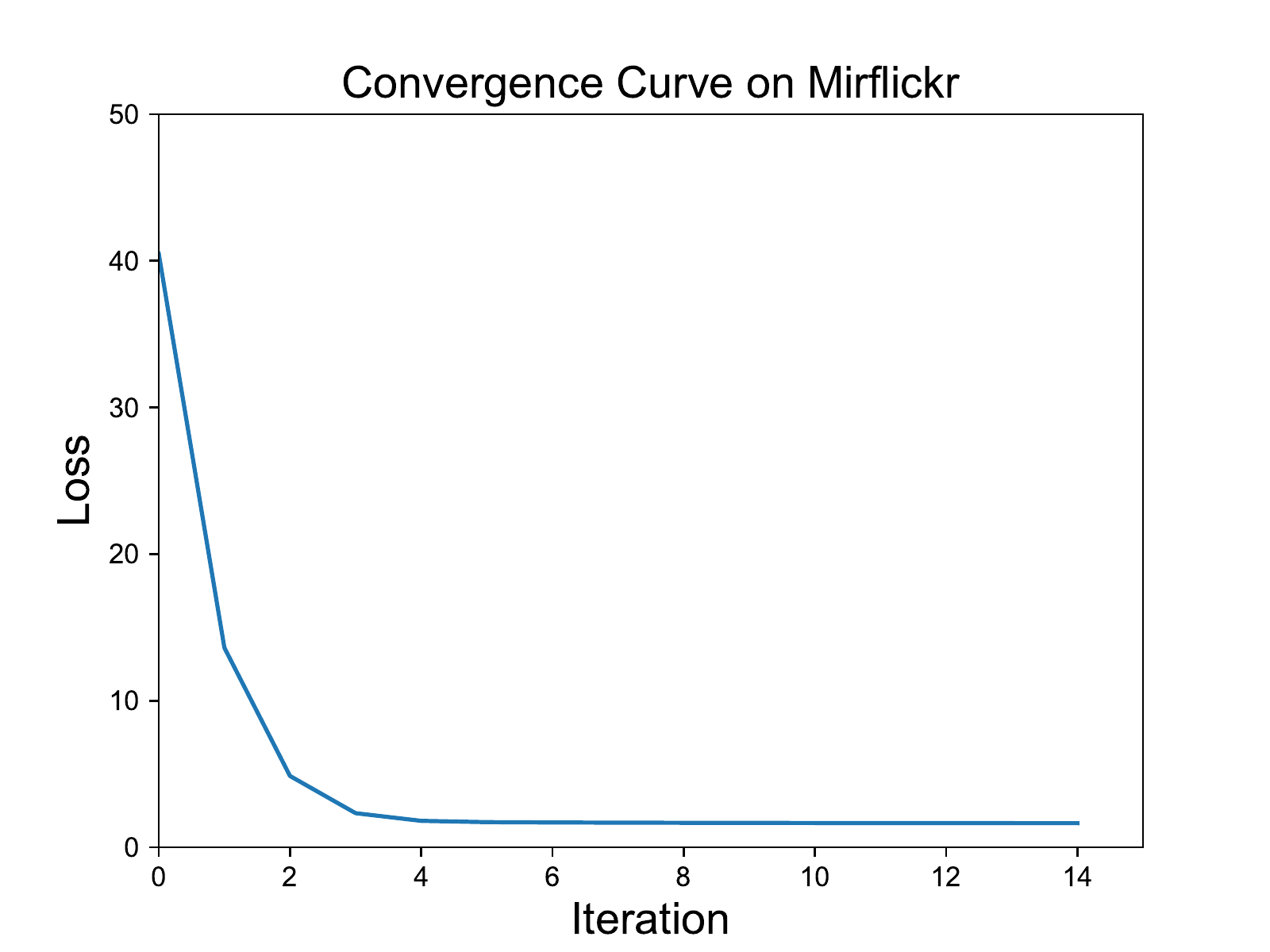} }
	\subfigure[\label{fig:gf}][Soccer Player]{
		\includegraphics[scale=0.168]{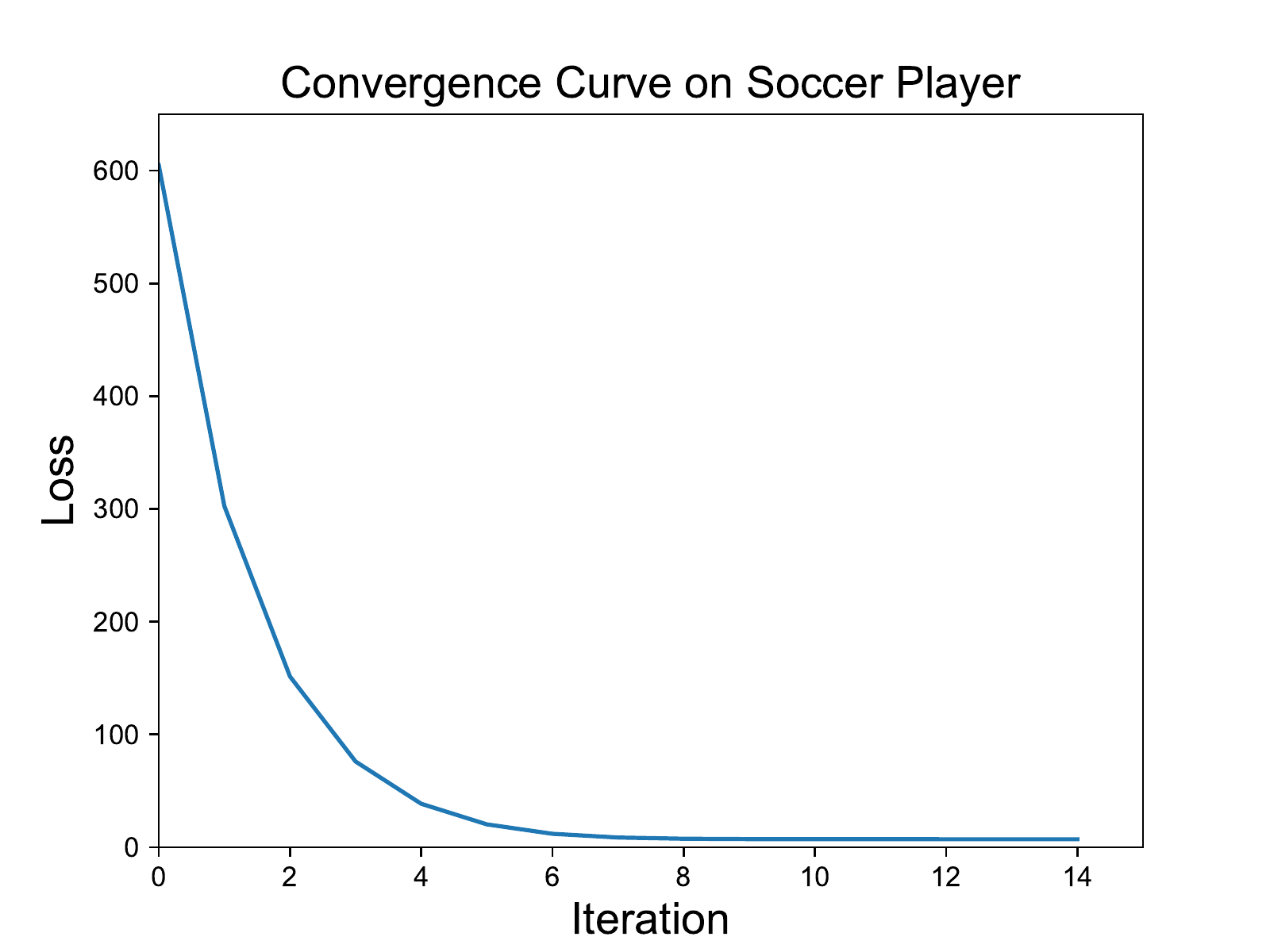} }
	\subfigure[\label{fig:gf}][Yahoo!News]{
		\includegraphics[scale=0.168]{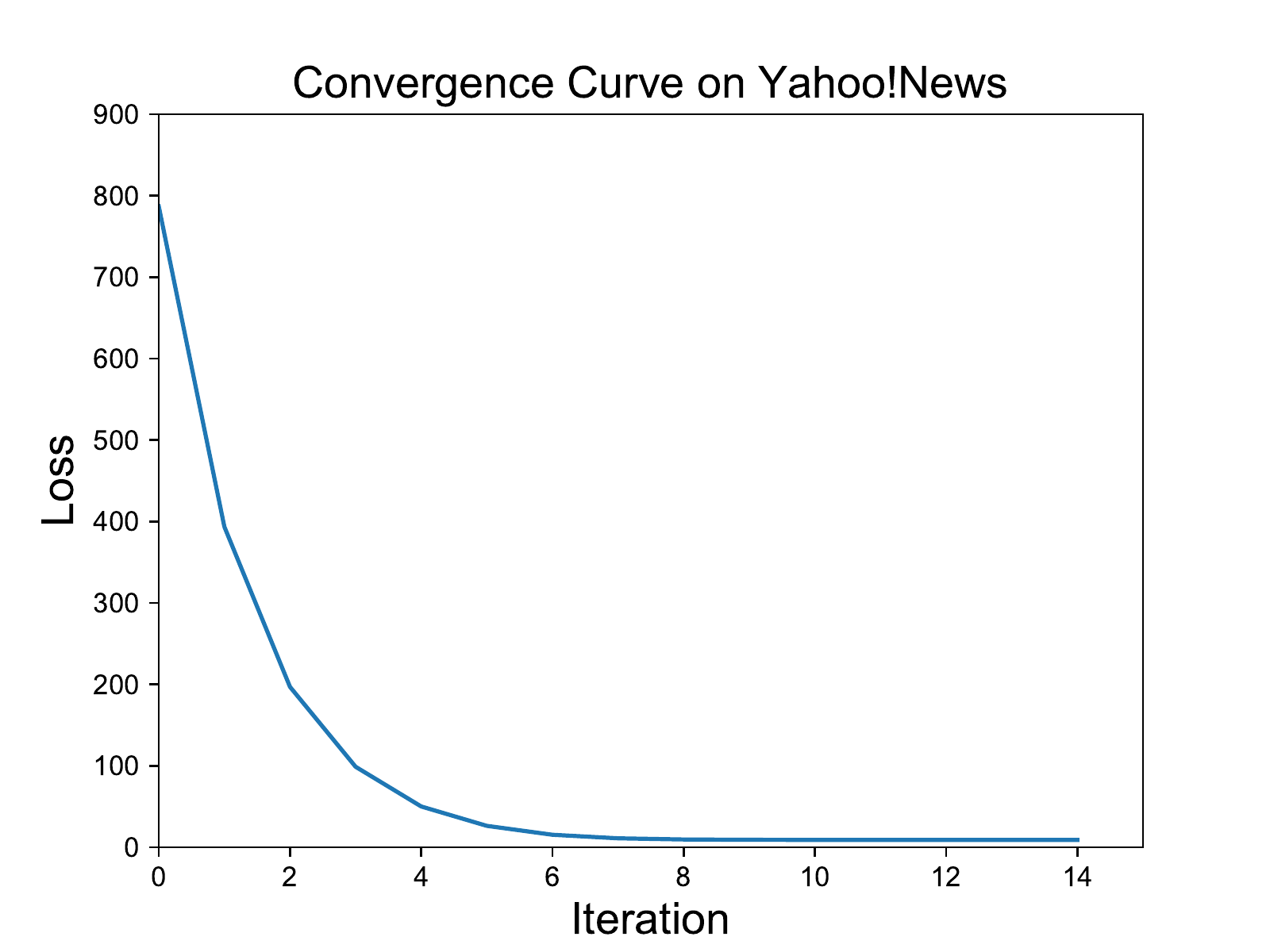} }
		
	\caption{Sensitivity and convergence curves of PL-CL.}
	\label{fig sensitive} 
\end{figure*}

\subsection{Performance on Real-World Data Sets}

We evaluated the performance of our method and the compared ones on 6 real-world data sets, whose details are summarized in Table \ref{tab:real-world data character}. For facial age estimation such as FG-NET \cite{panis2016overviewFg-net}, human faces are represented as instances while the ages annotated by crowd-sourcing labelers are considered as candidate labels. For automatic face naming, i.e., Lost \cite{cour2009learningLost}, Soccer Player \cite{zeng2013learningSoccerplayer} and Yahoo!News \cite{guillaumin2010multipleYahoonews}, each face cropped from an image or a video frame is taken as an instance with the names extracted from the corresponding captions or subtitles as its candidate labels. For object classification task like MSRCv2 \cite{liu2012conditionalMSRCv2}, image segments are considered as instances with objects appearing in the same image as candidate labels. For web image classification task such as Mirflickr \cite{huiskes2008mirMirflickr}, each image is presented as an instance with the tags extracted from the web page being candidate labels.

As the average candidate labels of FG-NET is very large, which may result in quite low classification accuracy. Following \cite{zhang2016partialPLLEAF}, we further employed mean absolute error (MAE) as the evaluation criteria. Specifically, for FG-NET(MAE3)/(MAE5), test examples are considered to be correctly classified if the error between predicted age and true age is no more than 3/5 years.

Table \ref{tab:real world statistic} shows the classification accuracy with standard deviation of each approach on the real-world data sets, where we can observe that
\begin{itemize}
    \item PL-CL significantly outperforms PL-AGGD, LALO, PLDA and PL-KNN on all the real-world data sets. Compared with two graph-based approaches PL-AGGD and LALO, PL-CL achieves outstanding performance, which validates the effectiveness of the proposed complementary classifier.
    \item IPAL performs well on MSRCv2 and Yahoo!News, while it is not apt to the other data sets. However, PL-CL performs well on all the data sets, which shows the robustness and effectiveness of PL-CL.
    \item In a nutshell, PL-CL achieves significant superior accuracy in 95.8\% cases according to the pairwise $t$-test at 0.05 significance level.
\end{itemize}

\subsection{Performance on Controlled UCI Data Sets}
Table \ref{tab:uci data character} summarizes the characteristics of four UCI data sets. Following the widely-used controlling protocol in PLL \cite{zeng2013learningSoccerplayer}, \cite{liu2012conditional23462372}, a synthetic partial label data set can be generated from the UCI data set, which is controlled by three parameters $p$, $r$ and $\epsilon$. Specifically, $p$ controls the proportion of the training instances with partial labels, $r$ denotes the number of false positive labels residing in the candidate label set, and $\epsilon$ stands for the co-occurring probability of one candidate label and the ground-truth label.

With fixed $p=1$ and $r=1$, the classification accuracy of all the methods when $\epsilon$ varies from 0.1 to 0.7 with step size 0.1 is shown in Fig. \ref{fig performance uci}(a). A specific label was chosen to co-occur with the ground-truth label with probability $\epsilon$, and the rest of the labels were randomly chosen with probability $(1-\epsilon)/(l-2)$. Fig. \ref{fig performance uci}(b)-(d) show the classification accuracy with varying $p$ from 0.1 to 0.7 when $r=1,2,3$, respectively. We randomly selected proportion $p$ of the training samples as partial label examples with randomly chosen $r$ labels as the corresponding false positive labels, i.e., the size of the candidate labels of each instance is $r+1$\footnote{Under this setting, as the false positive labels are randomly chosen, $\epsilon$ is set to $\frac{1}{l-1}$.}. Table \ref{tab:uci data wintieloss} summarizes the detailed win/tie/loss counts between PL-CL and other methods according to the pairwise $t$-test with a significance level of 0.05. From the Fig. \ref{fig performance uci} and Table \ref{tab:uci data wintieloss} we can conclue that:

\begin{itemize}
    
    \item With increasing $p$ and $\epsilon$, the classification accuracy drops gradually in general for all the methods, as a larger $p$ or a larger $\epsilon$ means a more difficult PLL task. Even though, our approach also performs better than the compared methods. 
    \item As the number of candidate labels increases, the gaps between PL-CL and others become more obvious, which means with more candidate labels, the effect of complementary classifier is more prominent.
    \item PL-CL clearly outperforms other methods with all the settings in general, i.e., PL-CL outperforms other compared methods significantly in 92.1\% cases and achieves at least comparable performance in 7.9\% cases with the total of 672 cases. Moreover, PL-CL is never significantly outperformed by any compared approaches.

\end{itemize}

\subsection{Further Analysis}
\subsubsection{Disambiguation Ability}
Transductive accuracy evaluates the classification accuracy of a PLL method on the training examples, which could reflect the ability of label disambiguation. Table \ref{tab:real world statistic transductive} shows the transductive accuracy of all the methods on all the real-world data sets, where we can find that PL-CL outperforms other compared methods in 89.5\% cases and is at least comparable to them in 6.3\% cases according to the pairwise $t$-test, which validates the disambiguation ability of PL-CL.

\subsubsection{Ablation Study} 

In order to validate the effectiveness of the complementary classifier and the adaptive graph, ablation studies are conducted and the results are shown in Table \ref{tab:real world further}. It can be observed from the Table \ref{tab:real world further} that the complementary classifier is quite useful when comparing the performance of the second and forth rows or the third and fifth rows on classification accuracy, as the performance of the method with complementary classifier is superior to that without the complementary classifier. Additionally, when we compare the last two rows of the table, we can find that the model with the graph structure achieves superior performance than that without the graph. In general, PL-CL outperforms its degenerated versions in all cases w.r.t. both classification accuracy and transductive accuracy, which proves that both the complementary classifier and graph structure play significant roles in our model. Besides, the use of kernel can also improve the performance of PL-CL, which can be observed when comparing the results in the first and second rows. It is worth noting that although kernel contributes to the improvement of the performance, the main improvement of PL-CL comes from complementary classifier and the graph.

\subsubsection{Sensitive and Convergence Analysis}

Figs. \ref{fig sensitive}(a)-(e) illustrate the sensitivity of PL-CL w.r.t. $\alpha$, $\beta$, $\mu$, $\gamma$ and $\lambda$ on Lost, where we can find that PL-CL performs quite stably as the values of the hyper-parameter change within a reasonable wide range. Additionally, Figs. \ref{fig sensitive}(f)-(k) show the convergence curves of PL-CL, where PL-CL converges fast within about 8 iterations on all the data sets, which is also a desirable property in practice.

\section{Conclusion and Future Work}
In this paper, a novel method PL-CL based on complementary classifier has been proposed to solve the PLL task. Different from previous works, we incorporated the information of complementary labels by inducing a complementary classifier, and use the complementary classifier to achieve candidate label disambiguation by a novel adversarial prior. Besides, PL-CL adopted an adaptive local graph shared by both the feature space and label space to assist disambiguation. Comprehensive experiments on four controlled UCI data sets as well as six real-world data sets validated the effectiveness of PL-CL and the usefulness of the proposed complementary classifier. In the future, we will investigate how to design deep-learning-based PLL models by leveraging the complementary classifier.

\bibliographystyle{ACMReferenceFormat}
\bibliography{samplebase}


\end{document}